\documentclass[letterpaper, 10 pt, journal, twoside]{IEEEtran}

\IEEEoverridecommandlockouts                              % This command is only needed if 
\usepackage{gensymb}

\usepackage{graphicx} % for pdf, bitmapped graphics files
\usepackage{amsmath} % assumes amsmath package installed

\usepackage{tabularx}

\usepackage{tabularray}

\usepackage[export]{adjustbox}

\usepackage{amsfonts} % mathbb

\usepackage{booktabs}
\usepackage{multirow}

\usepackage{hyperref}
\usepackage{svg}

\usepackage{subfigure}
\usepackage{color}
\usepackage{soul}

\begin{document}
\title{GroundGrid: LiDAR Point Cloud Ground Segmentation and Terrain Estimation}

% Paper headers
\markboth{IEEE Robotics and Automation Letters. Preprint Version. Accepted October, 2023}
{Steinke \MakeLowercase{\textit{et al.}}: GroundGrid: Point Cloud Ground Segmentation and Terrain Estimation} 
% Use only for final RAL version

\author{Nicolai Steinke$^{1}$, Daniel Goehring$^{1}$ and Ra\'{u}l Rojas$^{1}$% <-this % stops a space
\thanks{Manuscript received: June, 15, 2023; Revised October, 4, 2023; Accepted October, 28, 2023.}%Use only for final RAL version
\thanks{This paper was recommended for publication by Editor Pauline Pounds upon evaluation of the Associate Editor and Reviewers' comments.}%
\thanks{This work was supported by the German Federal Ministry for Digital and Transport under grant number 45AVF3001F.}% %Use only for final RAL version
\thanks{$^{1}$Dahlem Center for Machine Learning and Robotics (DCMLR), Department of Mathematics and Computer Science, Freie Universit\"{a}t Berlin, Germany, 
{\tt\small \{nicolai.steinke | daniel.goehring | raul.rojas\}@fu-berlin.de}}%
\thanks{Digital Object Identifier (DOI): see top of this page.}
}

©2024 IEEE.  Personal use of this material is permitted.  Permission from IEEE must be obtained for all other uses, in any current or future media, including reprinting/republishing this material for advertising or promotional purposes, creating new collective works, for resale or redistribution to servers or lists, or reuse of any copyrighted component of this work in other works.

The final published version is available on IEEE Xplore: \url{https://ieeexplore.ieee.org/document/10319084}\\
DOI: 10.1109/LRA.2023.3333233

\maketitle
%%%%%%%%%%%%%%%%%%%%%%%%%%%%%%%%%%%%%%%%%%%%%%%%%%%%%%%%%%%%%%%%%%%%%%%%%%%%%%%%
\begin{abstract}
The precise point cloud ground segmentation is a crucial prerequisite of virtually all perception tasks for LiDAR sensors in autonomous vehicles. Especially the clustering and extraction of objects from a point cloud usually relies on an accurate removal of ground points. The correct estimation of the surrounding terrain is important for aspects of the drivability of a surface, path planning, and obstacle prediction. In this article, we propose our system GroundGrid which relies on 2D elevation maps to solve the terrain estimation and point cloud ground segmentation problems. We evaluate the ground segmentation and terrain estimation performance of GroundGrid and compare it to other state-of-the-art methods using the SemanticKITTI dataset and a novel evaluation method relying on airborne LiDAR scanning. The results show that GroundGrid is capable of outperforming other state-of-the-art systems with an average IoU of 94.78\% while maintaining a high run-time performance of 171Hz. The source code is available at {\footnotesize \url{https://github.com/dcmlr/groundgrid}}
\end{abstract}
%%%%%%%%%%%%%%%%%%%%%%%%%%%%%%%%%%%%%%%%%%%%%%%%%%%%%%%%%%%%%%%%%%%%%%%%%%%%%%%%2
\section{INTRODUCTION}
\IEEEPARstart{L}{iDAR} sensors are widely used in the field of autonomous vehicles. Many applications for LiDAR sensors in this field require the removal of ground points. This applies especially to object detection and classification algorithms where the ground segmentation is often the first processing step that precedes higher perception functions, e. g., object detection and classification~\cite{li20}. With technological advances, the point density of the sensors rises continuously which makes scalable solutions necessary. As a related problem to ground segmentation, terrain estimation is the task of estimating the ground topology around the vehicle. Knowledge about the surrounding ground topology can be used to segment point clouds into ground and non-ground, infer drivable terrain, improve object movement prediction, and filter outlier points of the LiDAR sensor. In this article, we present GroundGrid, a LiDAR ground segmentation and terrain estimation system that outperforms other state-of-the-art methods while maintaining online performance. The method is sensor independent, deterministic, and easy to understand since it relies only upon the vertical point height variance and the outlier detection in order to detect the ground surface. We evaluate the ground segmentation performance on the well-known SemanticKITTI dataset~\cite{behley19} and compare it to other state-of-the-art methods. Additionally, we evaluate the terrain estimation performance using georeferenced Airborne Lidar Scanning (ALS) data as ground truth and compare the results with another method. 
%%%%%%%%%%%%%%%%%%%%%%%%%%%%%%%%%%%%%%%%%%%%%%%%%%%%%%%%%%%%%%%%%%%%%%%%%%%%%%%%
\section{RELATED WORK}\label{sec:related_work}
Since the ground segmentation of point cloud data is a fundamental task for LiDAR perception systems, especially for those in use on autonomous robots the research effort in this topic has been ongoing for many years~\cite{gomes23}.
The biggest challenges for accurate ground segmentation methods are the sparsity of the generated data, the fact that the point clouds are generally not spatially structured, and tight performance constraints to run online in the vehicle's on-board systems. To overcome these challenges, many approaches rely on the projection of the LiDAR point cloud to reduce the point cloud's dimensionality. Often the data is projected as a bird's eye view image onto a plane. Rasterizing this plane into a 2D grid structure and assigning the point height averages as cell values yields an elevation map. Algorithms relying on elevation maps were successfully used by many teams of the DARPA Grand Urban Challenge in 2007~\cite{kammel08}~\cite{montemerlo08}~\cite{umson08}. Elevation maps offer significant performance advantages because the calculations have only to be calculated for each cell of the map and not for each point in the point cloud. This fixed structure also enables the use of image processing algorithms and machine learning methods using convolutional neural networks~\cite{he22}~\cite{paigwar20}. Another popular way of projecting the data is in the form of a range image which results in a cylindrical projection for rotational laser scanners~\cite{bogoslavskyi16}~\cite{shen21}. Using this image representation allows classical image processing algorithms to be applied, e. g., 2D line regression algorithms~\cite{bogoslavskyi16} or methods relying on 2D image convolutions~\cite{shen21}. One disadvantage of these projection methods is the possible loss of information due to quantization errors~\cite{wu21}. Other works exploit the decline of the point density with the distance to the sensor. Himmelsbach et al.~\cite{himmelsbach10} discretized the point cloud anglewise so that each bin covered more area far away from the car than close to it. Others used polar elevation maps which also divide the space by angles so the cells cover more area the more distant they are to the sensor~\cite{cheng20}~\cite{lim21}. This way quantization errors can be minimized and the possible information loss reduced but this comes at the expense of a higher error in distant cells which also can include important information. Concerning the ground surface estimation, many methods rely on the assumption that the ground surface is relatively flat and continuous. Especially methods that use linear regression to fit lines or planes to the point cloud need these assumptions to hold~\cite{himmelsbach10}~\cite{douillard11}. In urban scenarios, the ground is often flat but not continuous since the ground level can abruptly change at road borders and pedestrian areas. In scenarios outside of the urban structure, the ground is often continuous but not flat since there are fewer human-made structures. In these cases, a singular plane can not accurately model the ground surface. Lim et al.~\cite{lim21} and the follow-up publication by Lee et al.~\cite{lee22} compensated for these problems by fitting multiple smaller planes. Himmelsbach et al.~\cite{himmelsbach10} and Cheng et al.~\cite{cheng20} split the area around the sensor into angular segments. All of these methods have in common that the fitted planes grow with the distance to the sensor. Since most sensors record fewer points per area in the distance the error concerning the recorded points remains constant. But the error for the true ground plane rises with the size of the planes. Since these algorithms are mostly evaluated on a per-point basis the bigger ground approximation error in the distance does not impact the results in these evaluations.\\
Recently methods utilizing artificial neural nets received more attention from researchers: Paigwar et al.~\cite{paigwar20} presented a convolutional neural network \textit{GndNet} that was trained on elevation maps and which infers the ground surface in order to segment point clouds. He et al.~\cite{he22} also split the area into segments of varying size similar to Lim et al.~\cite{lim21} and Lee et al.~\cite{lee22}. These segments were then processed with a PointNet~\cite{qi17} feature encoding block and subsequently fed into a classification module. With the classified sectors, the point cloud is segmented.
\begin{figure}[tpb]
 \centering
 \framebox{\parbox{0.8in}{\includegraphics[width=0.8in]{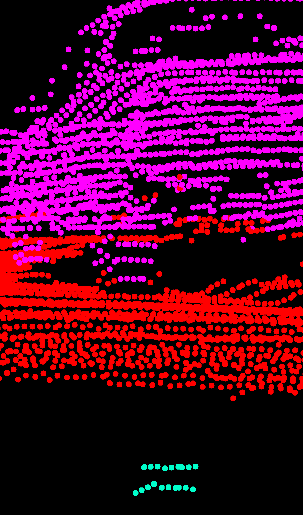}}}
 \caption{Visualization of outlier points (cyan) below the ground (red) due to a reflection on the car's body (purple). SemanticKITTI seq. 00 cloud 263.}
 \label{fig:outlier}
\end{figure}
A disadvantage of artificial neural nets is the need for large amounts of labeled data for training which often need to be generated manually. Paigwar et al.~\cite{paigwar20} and He et al.~\cite{he22} trained their networks on the SemanticKITTI dataset which consists of 11 publicly available manually labeled point cloud sequences. Since all the data was collected using the same sensor platform in similar locations around Karlsruhe, Germany, the dataset might be biased and the generalization capabilities of the outcomes are hampered.\\
Outlier removal in point clouds is rarely explicitly mentioned despite outliers are a common phenomenon - especially in urban scenarios where car chassis often reflect laser beams leading to incorrect measurements. Sequence 00 of the SemanticKITTI dataset is an example of an outlier prone scenario. Fig.~\ref{fig:outlier} is a visualization of common outlier points. The outlier points (cyan) are well below (approx. 1.5m) the true ground surface (red points) since two lasers have been reflected from the lower car body onto the ground and back to the LiDAR sensor. The sensor uses only the time of flight to determine the traveled distance of the light beam and does not seem to be able to detect the reflections of the beams. Hence the position of the points ends up being significantly lower than their true position. Spatial errors of this magnitude as well as the number of erroneous points poses a big challenge for ground estimation and segmentation systems. Lee~\cite{lee22} used two manually tuned thresholds to identify this kind of outlier points but they must be tuned to each sensor and mounting position individually and therefore this method does not provide a general solution to the problem.

%%%%%%%%%%%%%%%%%%%%%%%%%%%%%%%%%%%%%%%%%%%%%%%%%%%%%%%%%%%%%%%%%%%%%%%%%%%%%
\section{GROUND SEGMENTATION AND TERRAIN ESTIMATION}\label{sec:terrain_estimation}
\begin{figure}[tpb]
 \centering
 \includegraphics[width=2.4in]{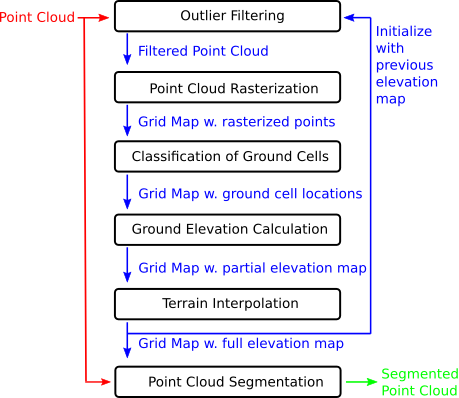}
 \caption{Overview of the proposed system. Input point cloud in red, intermediate results in blue, and output segmented point cloud in green.}
 \label{fig:overview}
\end{figure}

In this section, we describe GroundGrid in detail. Fig.~\ref{fig:overview} shows the general structure of the approach. The Input and output point clouds are depicted in red and green respectively and the intermediate results are in blue. The algorithm reuses the elevation maps of previous iterations. If no previous result is available the elevation map is initialized with the vehicle's elevation. First, an outlier filtering is executed, followed by the point cloud rasterization into a grid map. Subsequently, the grid map cells containing only ground points are classified and the ground level is calculated. Then the elevation map is generated and missing data is interpolated. Finally, the point cloud is segmented into ground points and non-ground points using the estimated elevation map.
\subsection{Outlier Filtering}\label{sec:outlier_filtering}
Reflected laser beams can yield points that lie well below the true ground level (see Fig.~\ref{fig:outlier}). These outliers lead to problems for the ground-level estimation. To remove outlier points, the intersection between the line connecting the sensor position with the point in question and the currently known elevation map is calculated. If the intersection shows that there is no line of sight between the sensor and the point, the point is classified as an outlier.
\subsection{Point Cloud Rasterization}
The remaining point cloud is rasterized into the grid map. During the rasterization process, several grid map layers are calculated: first of all the variance layer with the z-axis point coordinate variance for each cell. The other layers are the minimal, maximal, and average values of the z-coordinate (height) as well as the point count for each cell. The variance is calculated using Welford's online algorithm~\cite{welford1962} to avoid a second pass.
\subsection{Classification of Ground Cells} \label{sec:ground_cell_classification}
Cells containing exclusively ground points can be identified by using the variance information calculated in the previous step. The assumption is that traversable ground is more flat compared to obstacles and thus has a lower point height variance. In order to accommodate the sparsity of the point cloud data, the ground cell classification takes a cell's neighborhood into account. It is performed for patches of 9 ($3 \times 3$) to 25 ($5 \times 5$) cells depending on the distance to the sensor. If a cell does not contain enough points to calculate the variance, the average variance value of the patch is used. A cell in the center of a patch is classified as ground if the variance of the cell is below the variance threshold $t_v$. This threshold is scaled with the distance to the sensor to take the reduced accuracy of distant measurements into account:
\begin{equation} \label{eq:variance_threshold}
    t_{v} = d_{sf} \cdot d(p,o)
\end{equation}
where the distance scaling factor $d_{sf}$ is multiplied by the Euclidean planar distance $d$ of the point $p$ and the sensor origin $o$. The parameter $d_{sf}$ scales the distance to a variance threshold and is the same for all cells.
For a cell to be classified as ground it also must contain a sufficient number of points. Assuming the absence of obstacles, the LiDAR sensor used in the SemanticKITTI dataset produces a circle pattern for each laser on the ground. We approximate the expected point count per cell by calculating the angular distance covered by each cell on a circle and dividing it by the sensor's angular point distance:
\begin{equation} \label{eq:expected_points}
    n = \frac{\tan^{-1}(\frac{R}{d})}{d_{pv}}
\end{equation}
where $R$ is the cell resolution, $d$ is the distance to the LiDAR sensor and $d_{pv}$ is the angular distance of two points of this sensor. Cells that contain less than a predefined fraction (we used 0.25) of the expected points are skipped.
\subsection{Ground Elevation Calculation}
In this step, the ground elevation and a corresponding confidence value are calculated. It is again performed for patches of $9$ to $25$ cells (see sec.~\ref{sec:ground_cell_classification}). The ground elevation $h_{i,j}$ for the cell $i,j$ is calculated as follows:
\begin{equation}\label{eq:cur_groundlevel}
    h_{i,j} = \frac{\sum{(P_{i,j} \circ M_{i,j})}}{\sum{P_{i,j}}}
\end{equation}
where $P_{i,j}$ denotes the point count matrix at cell $i,j$ and $M_{i,j}$ the matrix containing the minimum height values.
The $\circ$-operator denotes the element-wise multiplication. The usage of point count weighted minimum height values as ground elevation minimizes the risk of a too high ground estimation which would result in under-segmentation. 
A ground cell's elevation estimation confidence $q_{i,j}$ at index ${i,j}$ is defined as
\begin{equation}\label{eq:cur_confidence}
    q_{i,j} = \frac{\sum{P_{i,j}}}{s}
\end{equation}
where $P_{i,j}$ is the point count matrix and $s$ the confidence point count scaling parameter. The resulting $q_{i,j}$ is clamped to $[0,1]$. 
Considering only ground cells, the elevation and confidence values are integrated with the existing values from previous observations as follows: The new elevation estimation $g_{i,j}$ is given by the confidence-weighted sum of the current and previous updates:
\begin{equation}\label{eq:groundlevel}
    g_{i,j} \gets \frac{q_{i,j} \cdot h_{i,j} + c_{i,j} \cdot g_{i,j}}{q_{i,j} + c_{i,j}}
\end{equation}
where $c_{i,j}$ and $g_{i,j}$ are values from previous updates while $h_{i,j}$ and $q_{i,j}$ are calculated according to~\ref{eq:cur_groundlevel} and~\ref{eq:cur_confidence}.
\begin{equation}\label{eq:confidence}
    c_{i,j} \gets \frac{1}{2}(\frac{q_{i,j}}{2} + c_{i,j})
\end{equation}
where $c_{i,j}$ denotes the existing confidence from previous steps for the cell $i,j$ and $q_{i,j}$ the confidence value for the current update. To give repeated ground detections a higher weight than a single detection with a high amount of points, the weight of $q_{i,j}$ is halved in this equation. We also calculate~\ref{eq:cur_groundlevel} for non-ground cells (variance $> t_v$). If $h_{i,j}$ is below the known ground level ($h_{i,j} < g_{i,j}$) we update it $g_{i,j} \gets h_{i,j}$, and we increase the confidence $c_{i,j}$ by a small amount: $c_{i,j} \gets \min(c_{i,j} + 0.1, 0.5)$ to reflect the addition of new information. This way under-segmentation is avoided for cells covered by obstacles.
\subsection{Terrain Interpolation}
The ground elevation is interpolated for cells where no ground points were detected. The interpolation is performed spirally from the sensor origin. The elevation for each cell is set to the confidence-weighted sum of the estimated ground elevation of the cell and the average of the cell's $3 \times 3$ neighborhood:
\begin{equation}\label{eq:inter_ground}
    g_{i,j} \gets (1 - c_{i,j}) \cdot \frac{\Sigma (C_{i,j} \circ G_{i,j})}{\Sigma C_{i,j}} + c_{i,j} \cdot g_{i,j}
\end{equation}
where $C_{i,j}$ is the $3 \times 3$ confidence matrix and $G_{i,j}$ the $3 \times 3$ ground elevation matrix centered at ${i,j}$. The $\circ$-symbol represents the element-wise multiplication. The confidence values are updated as follows:
\begin{equation}\label{eq:inter_confidence}
    c_{i,j} \gets c_{i,j} - \frac{c_{i,j}}{\theta}
\end{equation}
where $c_{i,j}$ is the confidence value at cell index $i,j$. The parameter $\theta$ is a constant decay factor that reduces the ground confidence over time.
\subsection{Point Cloud Segmentation}\label{sec:cloud_segmentation}
With the terrain elevation map, the point cloud can be segmented. We compare two threshold parameters $h_g$ and $h_o$ with each point's z-height above the corresponding cell's elevation for the segmentation. $h_g$ is for segmenting areas classified as ground while $h_o$ is for segmenting areas classified as obstacles. We set $h_o < h_g$ because the aim is to segment all of the obstacle's points correctly while avoiding over-segmentation in sloped ground areas.

%%%%%%%%%%%%%%%%%%%%%%%%%%%%%%%%%%%%%%%%%%%%%%%%%%%%%%%%%%%%%%%%%%%%%%%%%%%%%
\section{EXPERIMENTS AND EVALUATION}\label{sec:experiment_analysis}
\begin{figure*}[ht!]
  \centering
    \begin{tblr}{
    colspec = {lccc}, rowspec = {cccccc},
    }
    Ground Truth~\cite{behley19} &
   \begin{subfigure}
   \centering
   \includegraphics[height=1.2in,valign=c]{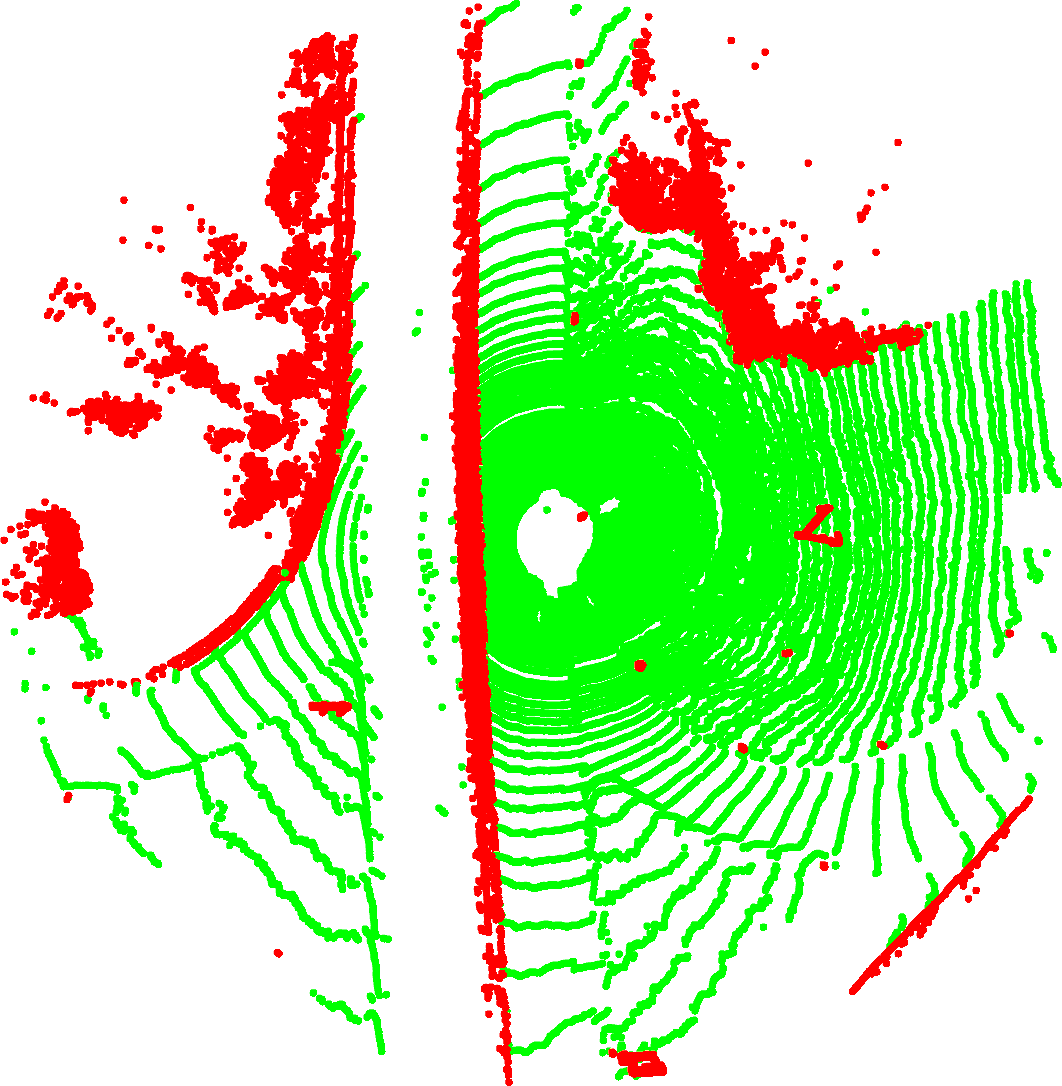}
   \end{subfigure} &
      \begin{subfigure}
   \centering
   \includegraphics[height=1.2in,valign=c]{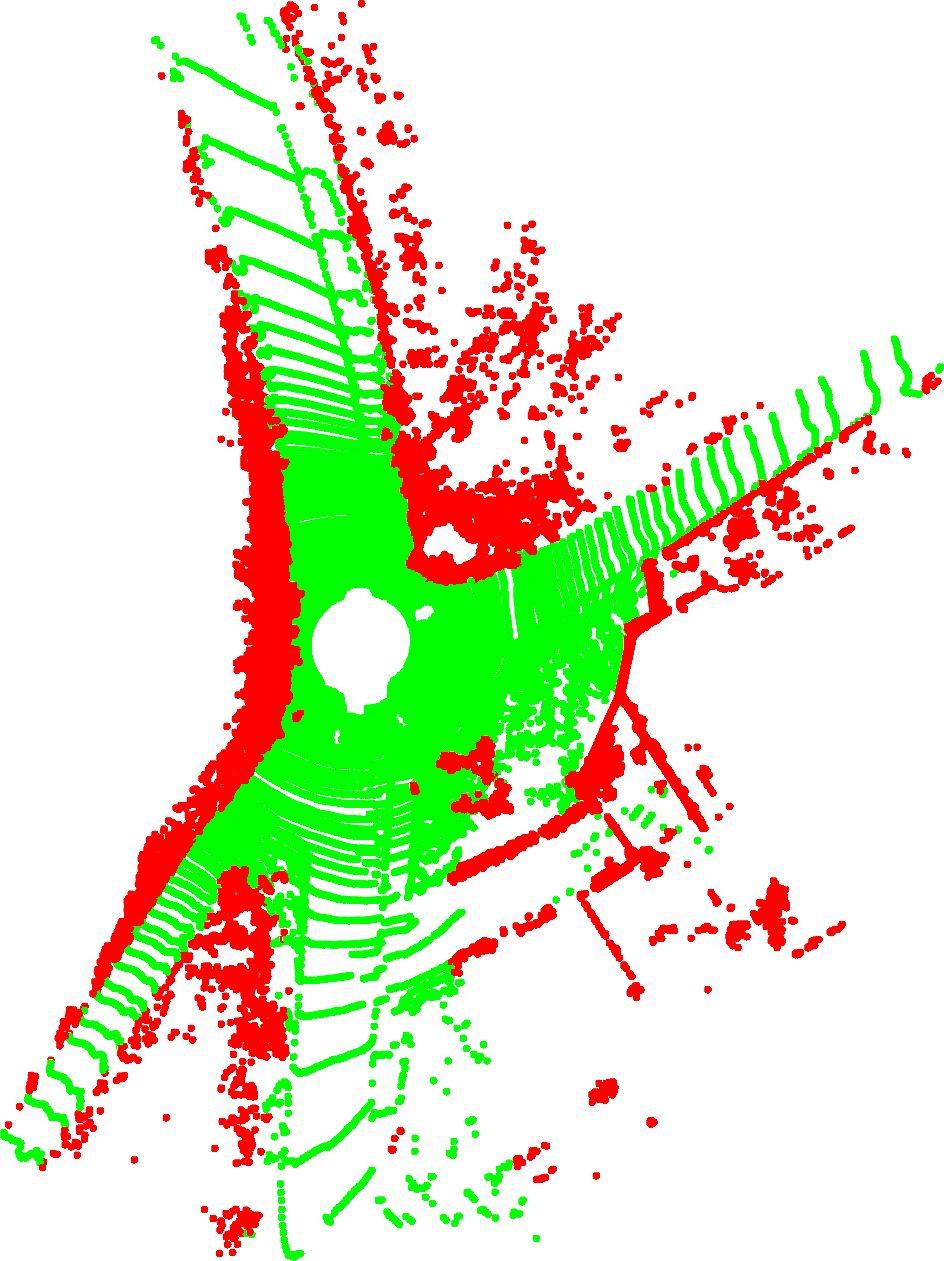}
   \end{subfigure} &
      \begin{subfigure}
   \centering
   \includegraphics[height=1.2in,valign=c]{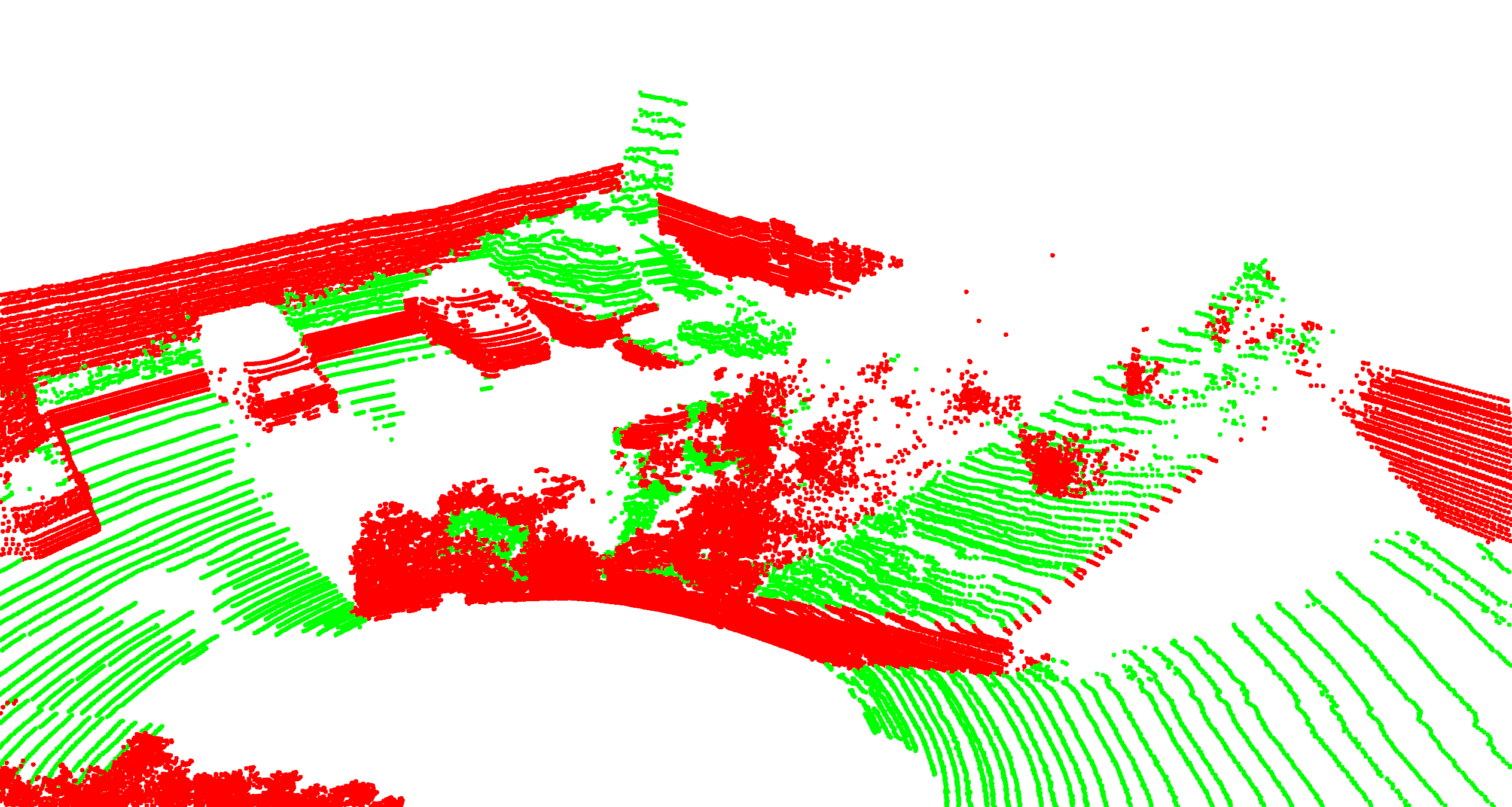}
   \end{subfigure}\\
   Patchwork++~\cite{lee22} &
      \begin{subfigure}
   \centering
   \hfill
   \includegraphics[height=1.2in,valign=c]{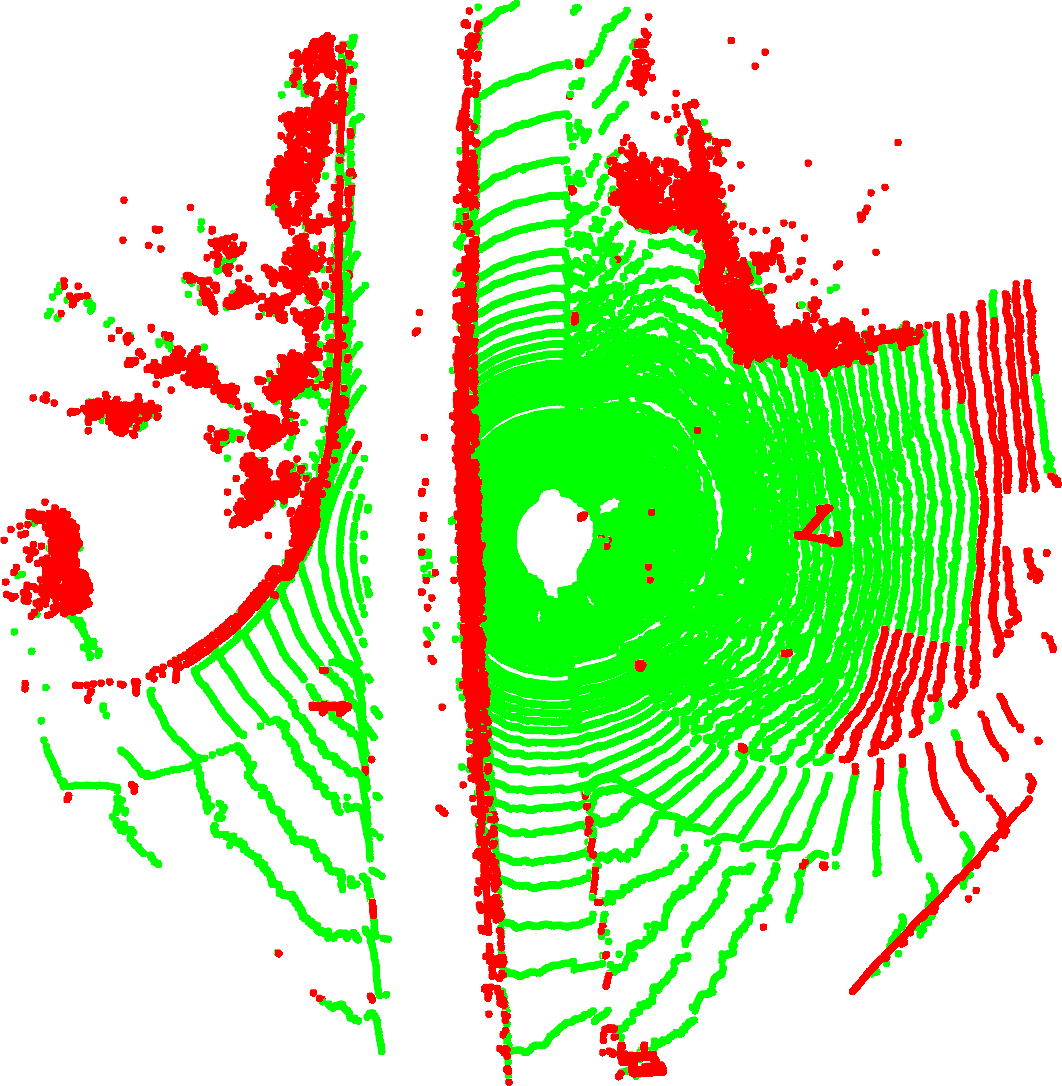}
   \end{subfigure}&
      \hfill
      \begin{subfigure}
   \centering
   \includegraphics[height=1.2in,valign=c]{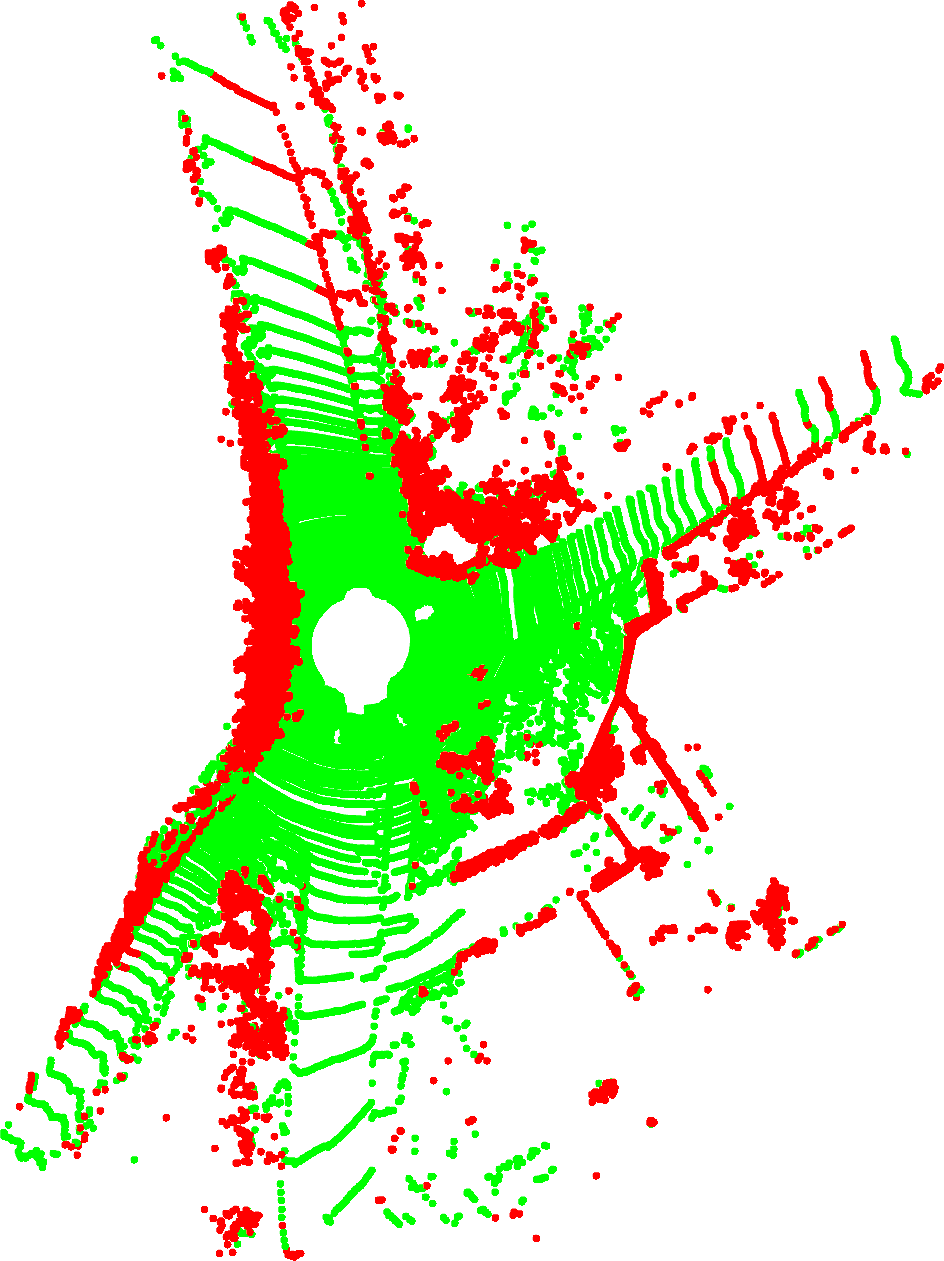}
   \end{subfigure}&
      \hfill
      \begin{subfigure}
   \centering
   \includegraphics[height=1.2in,valign=c]{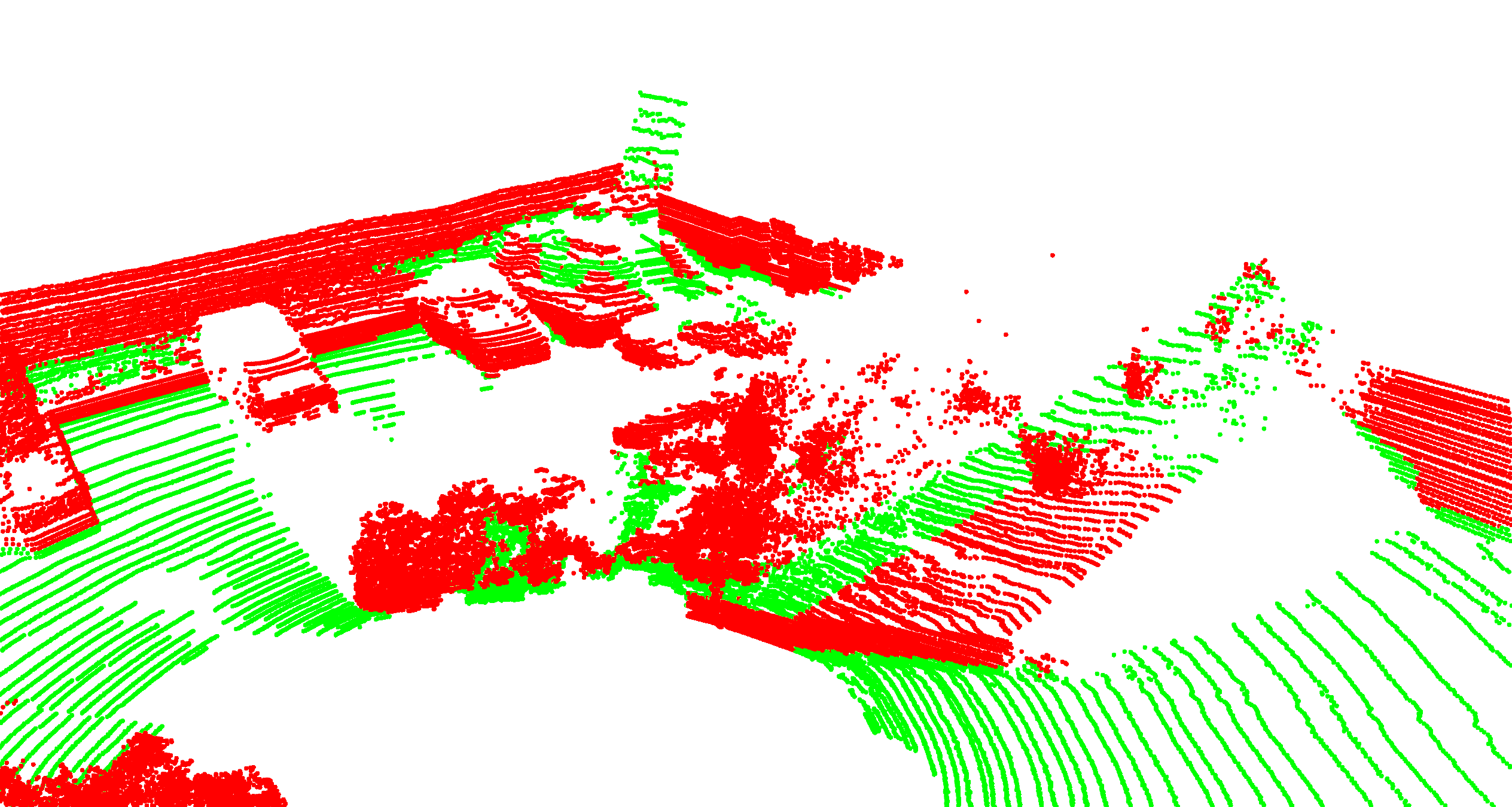}
   \end{subfigure}\\
  GndNet~\cite{paigwar20} &
      \begin{subfigure}
   \centering
   \hfill
   \includegraphics[height=1.2in,valign=c]{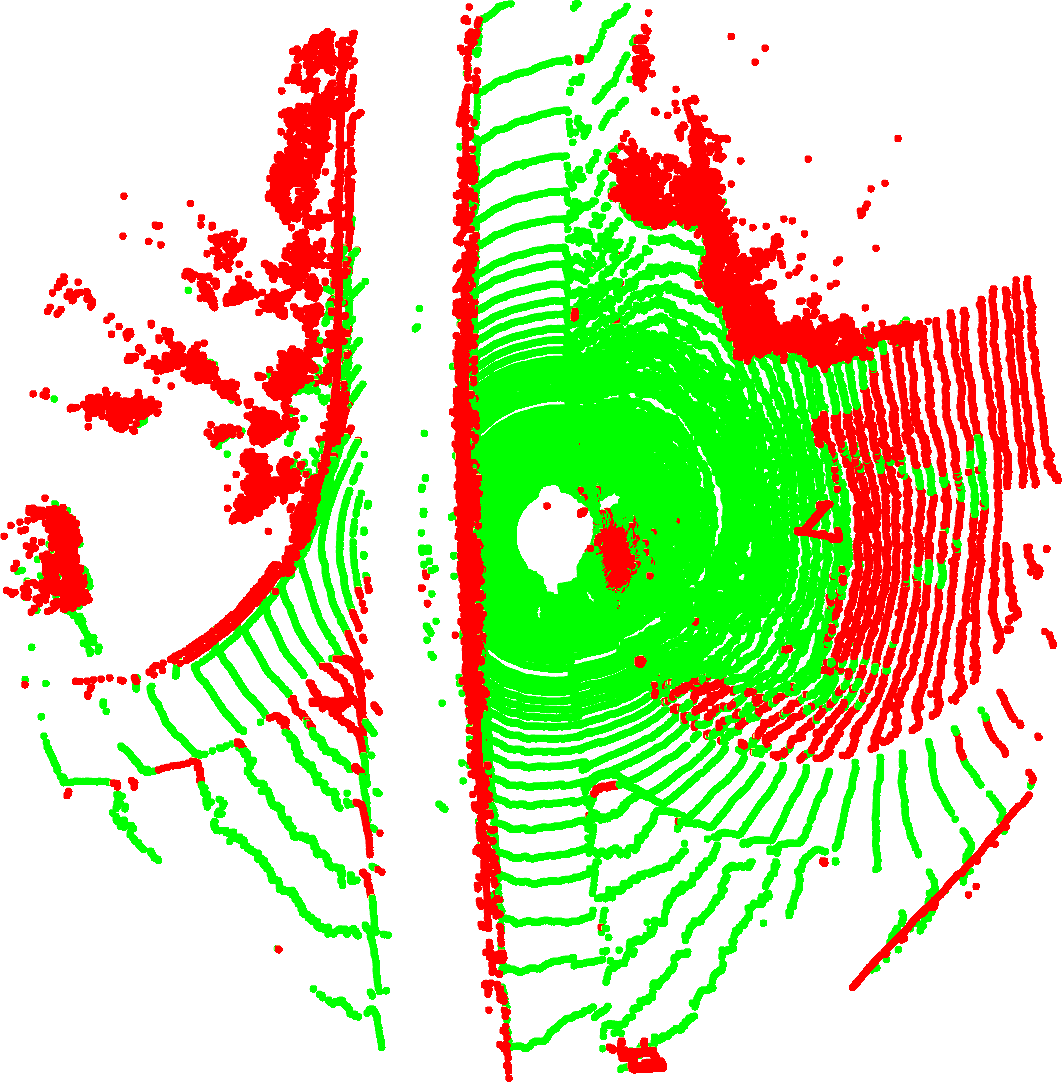}
   \end{subfigure}&
         \hfill
      \begin{subfigure}
   \centering
   \includegraphics[height=1.2in,valign=c]{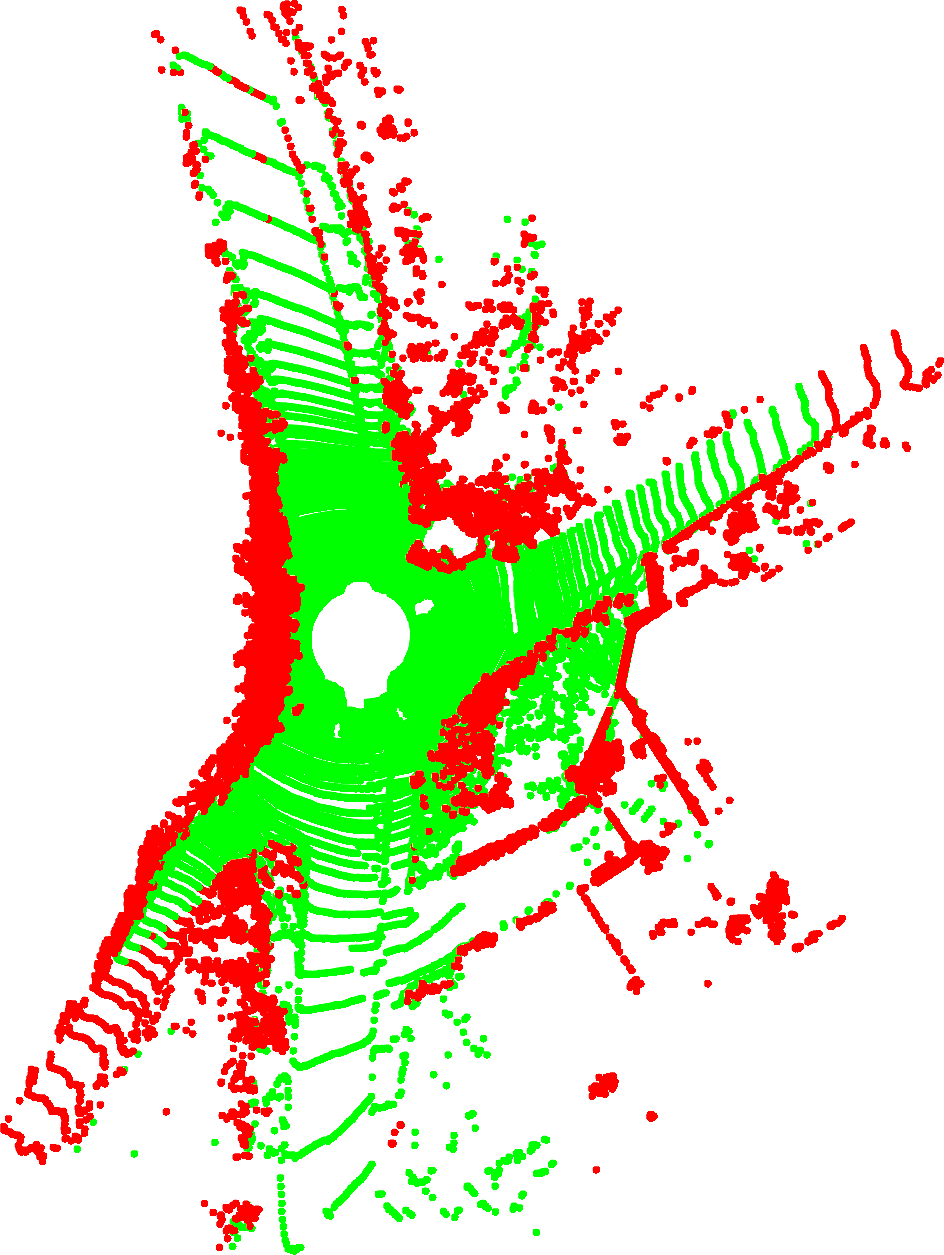}
   \end{subfigure}&
         \hfill
      \begin{subfigure}
   \centering
   \includegraphics[height=1.2in,valign=c]{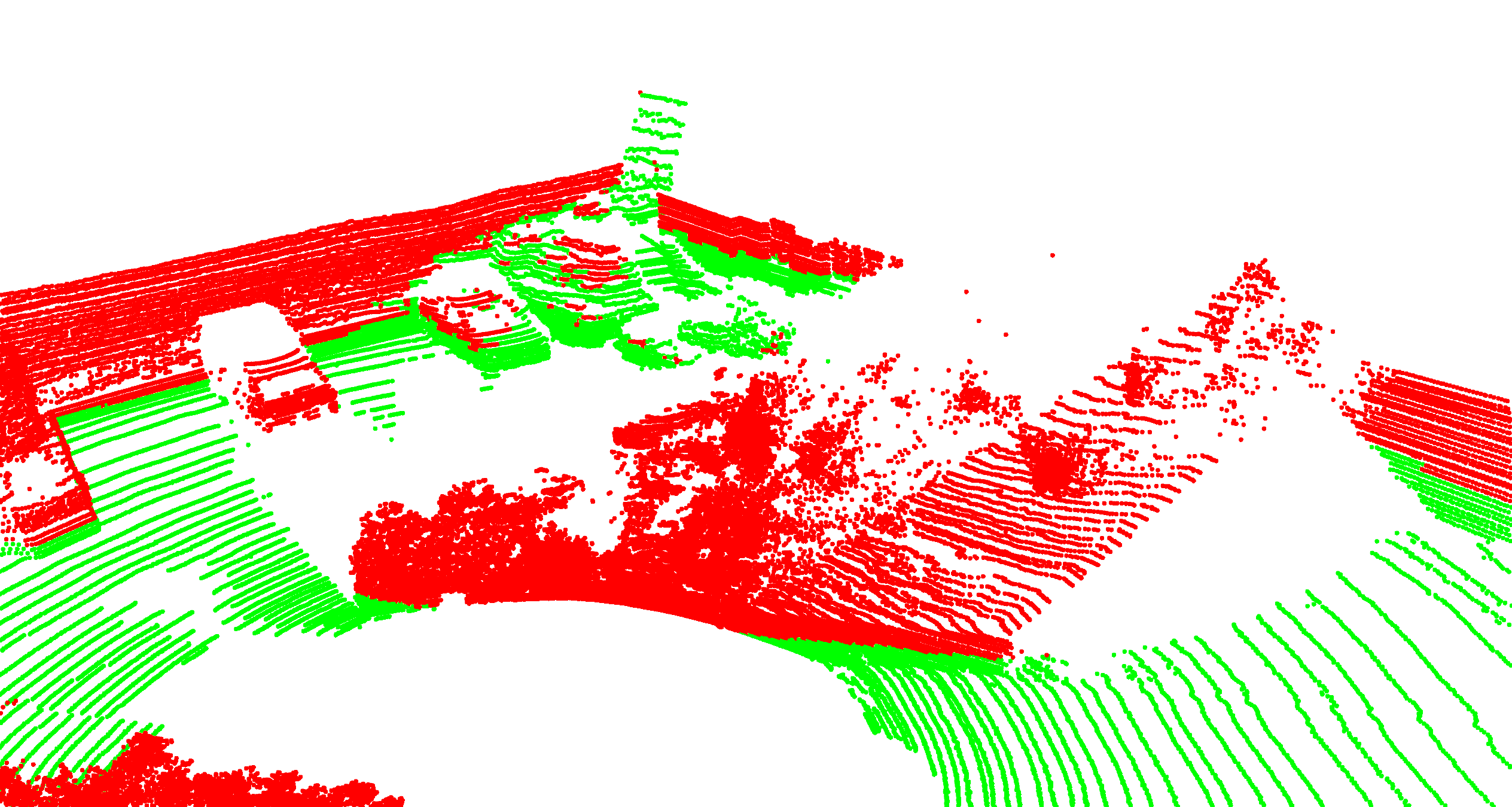}
   \end{subfigure}\\
   JPC~\cite{shen21} &
      \begin{subfigure}
   \centering
   \hfill
   \includegraphics[height=1.2in,valign=c]{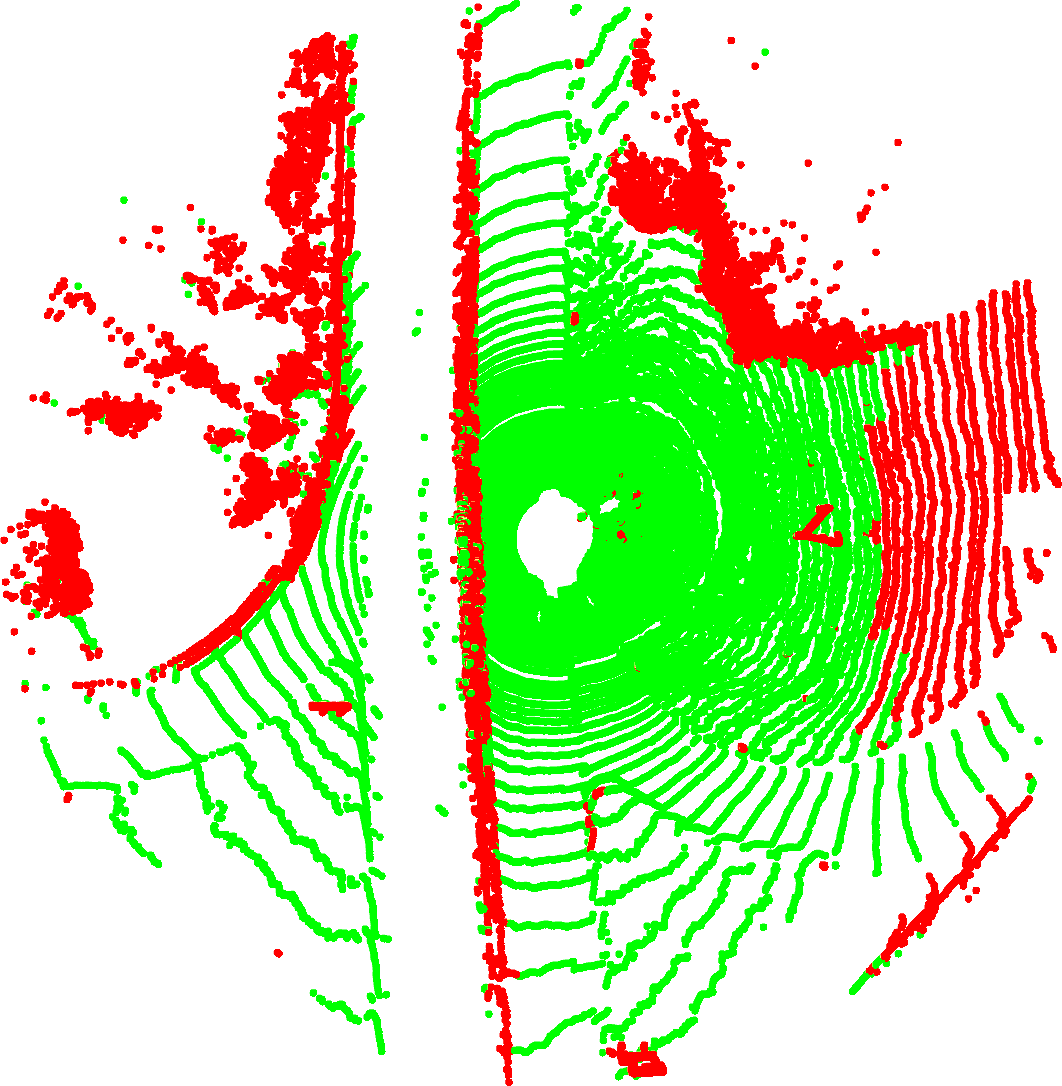}
   \end{subfigure}&
         \hfill
      \begin{subfigure}
   \centering
   \includegraphics[height=1.2in,valign=c]{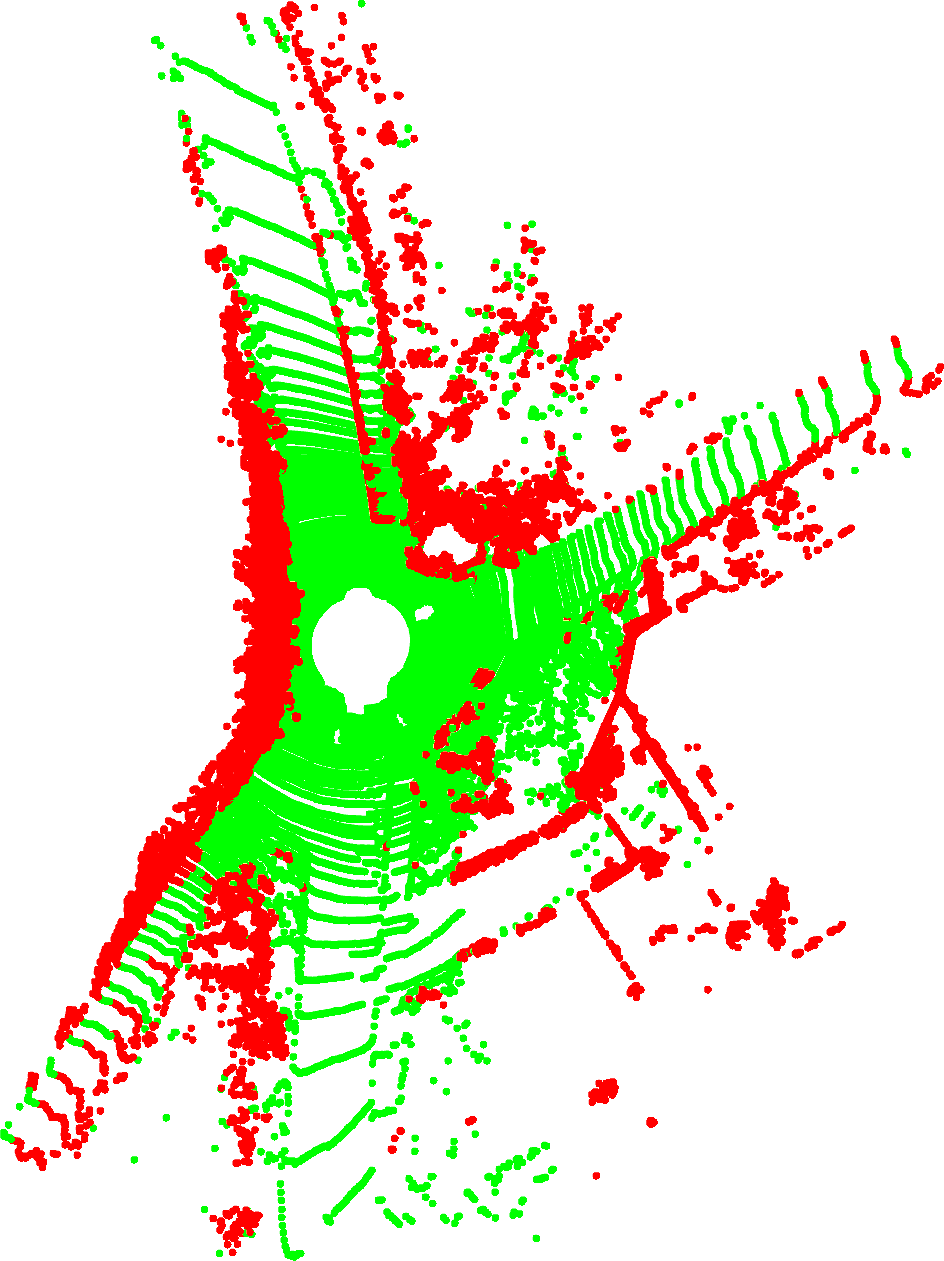}
   \end{subfigure}&
         \hfill
      \begin{subfigure}
   \centering
   \includegraphics[height=1.2in,valign=c]{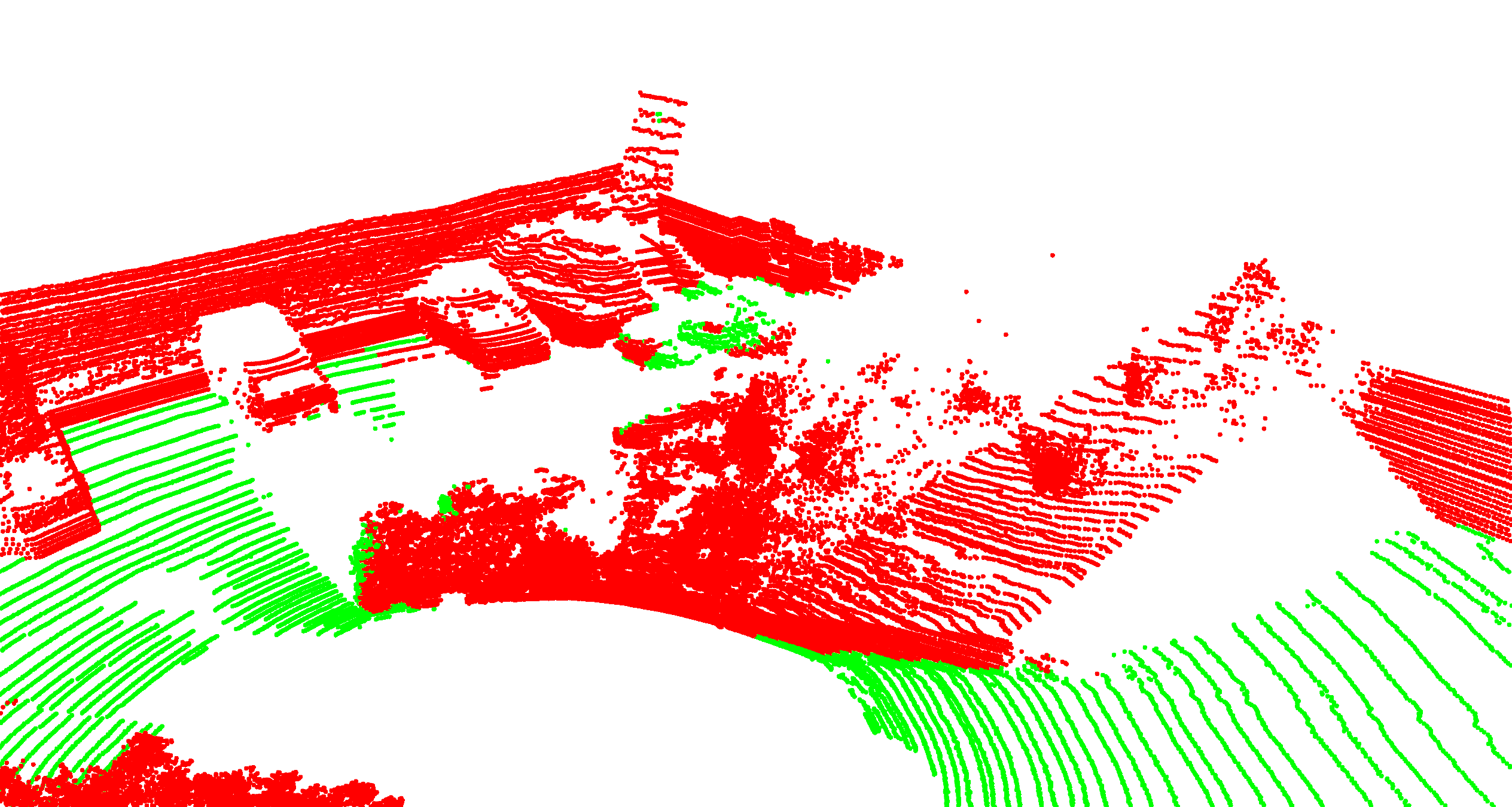}
   \end{subfigure}\\
   Ours &
      \begin{subfigure}
   \centering
   \hfill
   \includegraphics[height=1.2in,valign=c]{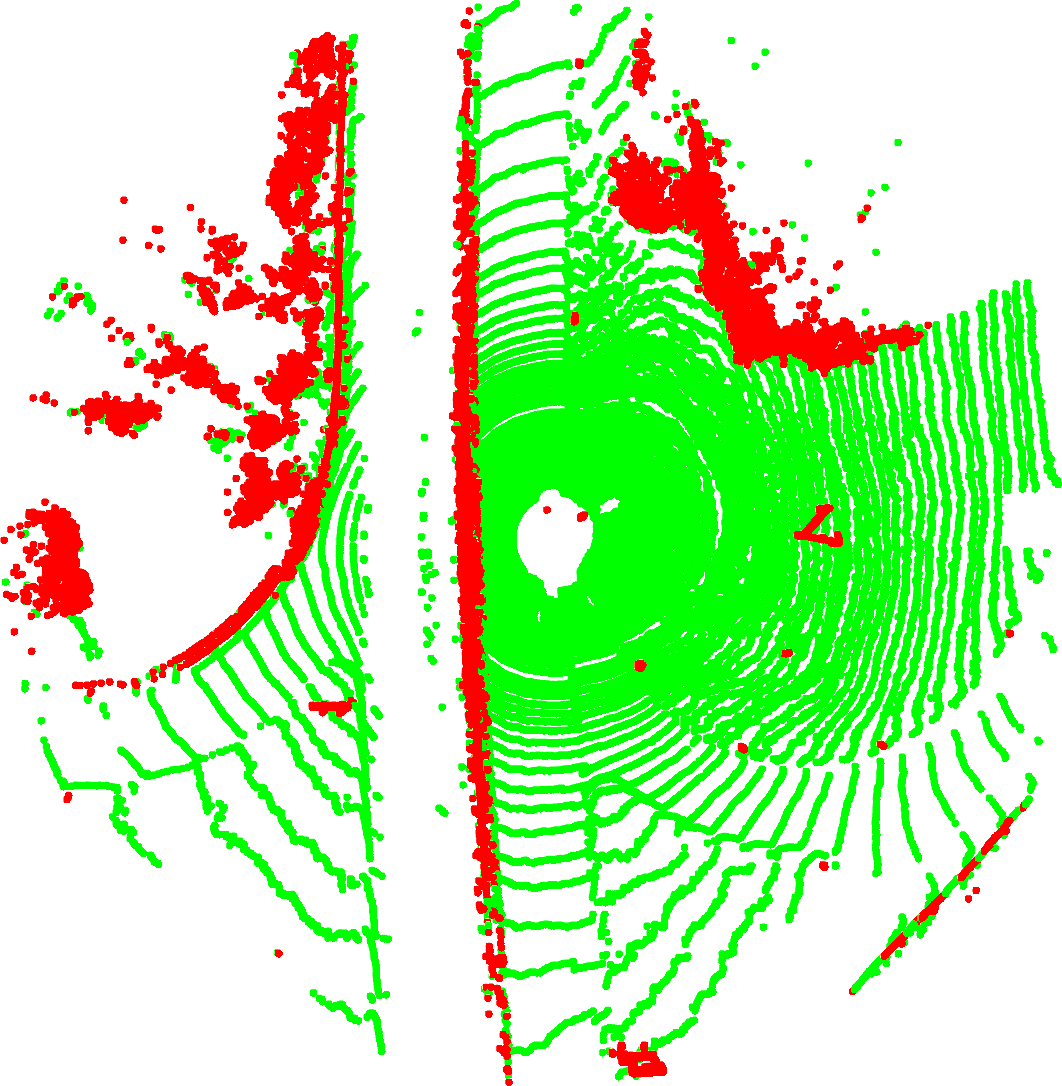}
   \end{subfigure}&
         \hfill
      \begin{subfigure}
   \centering
   \includegraphics[height=1.2in,valign=c]{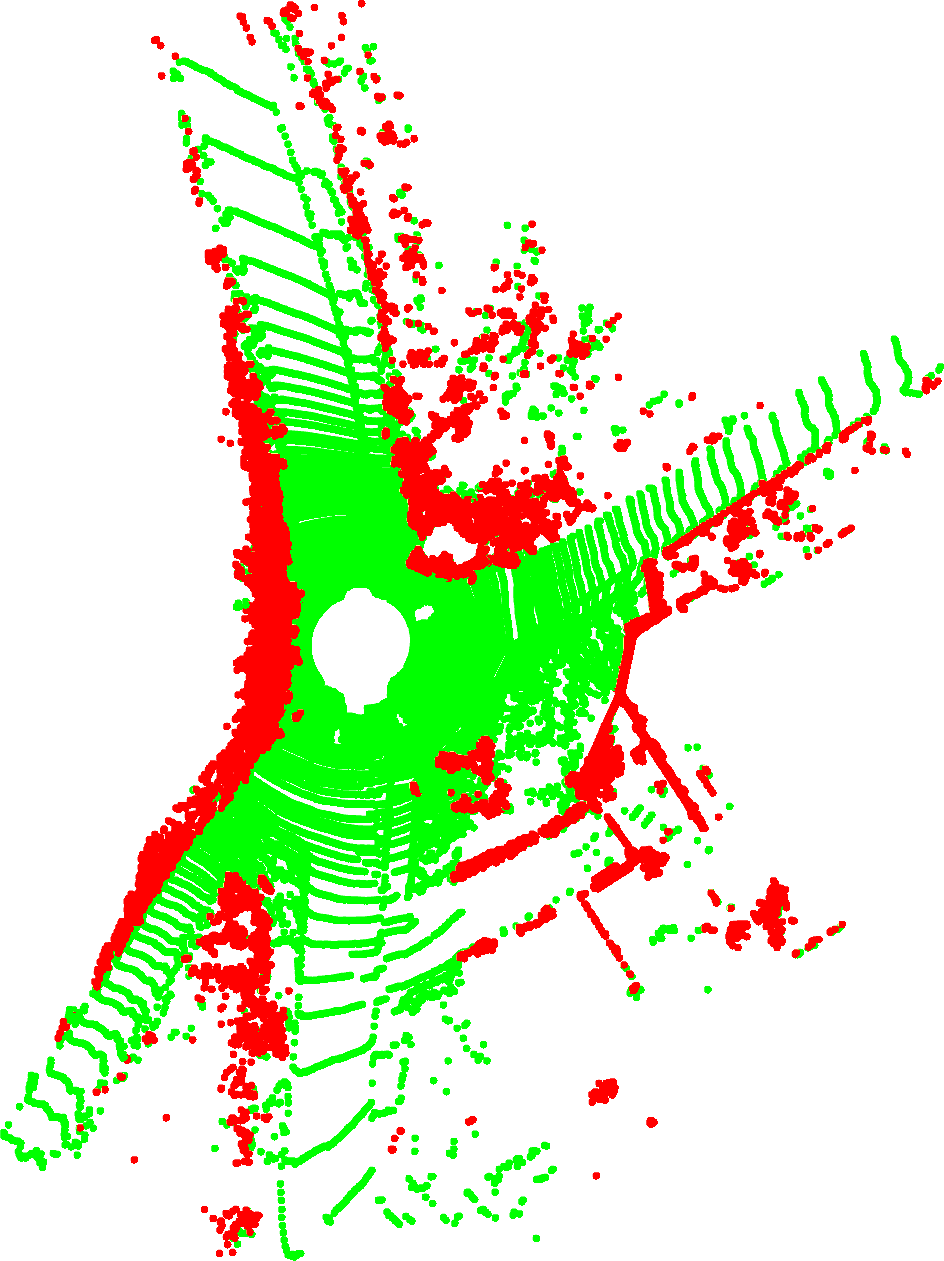}
   \end{subfigure}&
         \hfill
      \begin{subfigure}
   \centering
   \includegraphics[height=1.2in,valign=c]{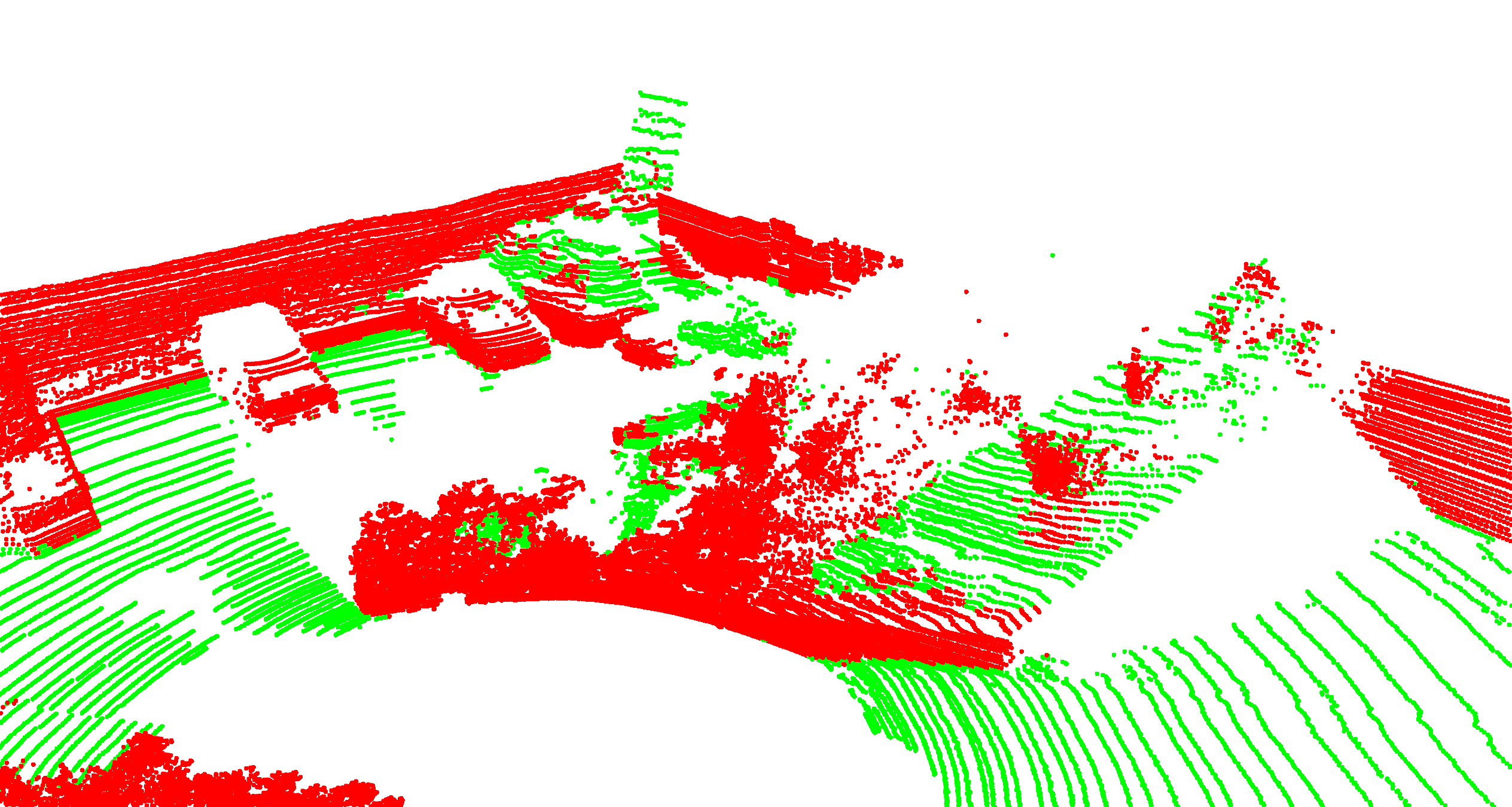}
   \end{subfigure}\\
   & a) & b) & c)\\
   \end{tblr}
   \caption{Ground segmentation qualitative evaluation (ground points in green / obstacle points in red)}
   \label{tab:qualitative_semkitti}
\end{figure*}
In this section, we present a qualitative and quantitative evaluation of the ground segmentation and terrain estimation results of the proposed system. We compare the ground segmentation results using the SemantikKITTI dataset to the state-of-the-art methods \textit{Patchwork++}~\cite{lee22}, \textit{Jump Point Convolution (JPC)}~\cite{shen21} and \textit{GndNet}~\cite{paigwar20} leveraging open source implementations\footnote{\url{https://github.com/url-kaist/patchwork-plusplus}}\footnote{\url{https://github.com/anshulpaigwar/GndNet}}\footnote{\url{https://github.com/wangx1996/Fast-Ground-Segmentation-Based-on-JPC}}. We evaluate the terrain estimation capabilities by using self-acquired data and a public georeferenced Airbourne Lidar Scanning (ALS) dataset~\cite{fis}.
\subsection{Parameters}
\begin{table*}[t]
  \centering
    \vspace{.5em}
   \caption{Ground segmentation accuracy evaluation}
   \begin{tabular}{l||ccccccccccc|c}
       seq. & {00} & {01} & {02} & {03} & {04} & {05} & {06} & {07} & {08} & {09} & {10} & {average}\\
       \midrule
       & & & & & & Precision\\
       \midrule
     Patchwork++ & 94.99 & \textbf{98.27} & 95.96 & 96.81 & 98.18 & 92.65 & 97.86 & 93.29 & 96.97 & 96.06 & 92.81 & 95.80\\
     GndNet & 92.40 & 96.54 & 93.74 & 95.60 & 97.30 & 89.58 & 96.15 & 90.09 & 95.09 & 93.81 & 88.34 & 93.51\\
     JPC & \textbf{96.78} & 97.97 & \textbf{97.50} & \textbf{98.09} & 99.01 & 94.03 & \textbf{97.96} & \textbf{95.65} & \textbf{97.97} & \textbf{97.64} & 95.27 & \textbf{97.08}\\
     Ours & 96.05 & 98.01 & 97.36 & 97.96 & \textbf{99.08} & \textbf{95.19} & 97.82 & 95.31 & 97.50 & 97.25 & \textbf{95.38} & 96.99\\
       \midrule
       & & & & & & Recall\\
       \midrule
     Patchwork++ & 98.67 & 96.52 & 97.20 & \textbf{98.17} & 97.21 & 98.13 & 97.39 & 98.42 & 97.41 & 96.45 & \textbf{95.93} & 97.41\\
     GndNet & \textbf{99.50} & \textbf{96.91} & 96.94 & 96.68 & \textbf{99.06} & \textbf{98.69} & \textbf{99.00} & \textbf{99.44} & \textbf{98.74} & 96.14 & 93.60 & \textbf{97.70}\\
     JPC & 97.20 & 95.46 & 93.72 & 94.86 & 96.91 & 95.64 & 96.23 & 96.53 & 95.13 & 92.66 & 88.47 & 94.97\\
     Ours & 98.70 & 96.17 & \textbf{97.71} & 97.95 & 97.85 & 98.13 & 98.38 & 98.72 & 97.79 & \textbf{96.91} & 95.90 & 97.65\\
       \midrule
       & & & & & & F1\\
       \midrule
     Patchwork++ & 96.80 & \textbf{97.39} & 96.58 & 97.49 & 97.69 & 95.31 & 97.63 & 95.79 & 97.19 & 96.25 & 94.35 & 96.59\\
     GndNet & 95.82 & 96.72 & 95.31 & 96.14 & 98.17 & 93.91 & 97.55 & 94.53 & 96.88 & 94.96 & 90.89 & 95.53\\
     JPC & 96.99 & 96.70 & 95.57 & 96.45 & 97.95 & 94.83 & 97.09 & 96.09 & 96.53 & 95.09 & 91.74 & 95.91\\
     Ours & \textbf{97.35} & 97.08 & \textbf{97.54} & \textbf{97.96} & \textbf{98.46} & \textbf{96.64} & \textbf{98.10} & \textbf{96.99} & \textbf{97.64} & \textbf{97.08} & \textbf{95.64} & \textbf{97.32}\\
       \midrule
       & & & & & & Accuracy\\
       \midrule
     Patchwork++ & 96.64 & \textbf{95.96} & 95.08 & 96.08 & 96.39 & 94.79 & 96.63 & 95.88 & 96.37 & 94.90 & 93.75 & 95.68\\
     GndNet & 95.53 & 94.88 & 93.20 & 93.99 & 97.10 & 93.10 & 96.46 & 94.52 & 95.91 & 93.08 & 89.81 & 94.33\\
     JPC & 96.89 & 94.93 & 93.80 & 94.59 & 96.81 & 94.37& 95.89 & 96.26 & 95.60 & 93.50 & 91.35 & 94.91\\
     Ours & \textbf{97.24} & 95.50 & \textbf{96.48} & \textbf{96.84} & \textbf{97.60} & \textbf{96.32} & \textbf{97.29} & \textbf{97.08} & \textbf{96.97} & \textbf{96.05} & \textbf{95.25} & \textbf{96.60}\\
       \midrule
       & & & & & & IoU\\
       \midrule
     Patchwork++ & 93.79 & \textbf{94.90} & 93.38 & 95.09 & 95.49 & 91.04 & 95.36 & 91.91 & 94.53 & 92.78 & 89.30 & 93.41\\
     GndNet & 91.97 & 93.65 & 91.04 & 92.55 & 96.41 & 88.52 & 95.22 & 89.63 & 93.94 & 90.41 & 83.30 & 91.51\\
     JPC & 94.15 & 93.61 & 91.52 & 93.14 & 95.98 & 90.16 & 94.34 & 92.47 & 93.29 & 90.64 & 84.75 & 92.19\\
     Ours & \textbf{94.84} & 94.33 & \textbf{95.19} & \textbf{96.00} & \textbf{96.97} & \textbf{93.49} & \textbf{96.27} & \textbf{94.15} & \textbf{95.40} & \textbf{94.33} & \textbf{91.64} & \textbf{94.78}\\
     \bottomrule
   \end{tabular}
   \label{tab:ground_segmentation}
 \end{table*}
In all experiments the parameters for our method were set as follows: We chose a grid map resolution of $R = 0.33m$  which offers a reasonable compromise between accuracy and computational performance. The variance threshold distance scaling factor was set to $d_{sf} = 10^{-5}$ (eq.~\ref{eq:variance_threshold}) and the minimum variance threshold to $t_{minv} = 5 \cdot d_{sf}$. The confidence decay was $\theta = 5$ (eq.~\ref{eq:inter_confidence}). The minimum point count relative to the expected point count per cell (eq.~\ref{eq:expected_points}) to perform the ground detection was $g_{minp} = 0.25$, cells with fewer points are skipped. The minimal ground confidence for the outlier detection (sum of a $5 \times 5$ matrix) was $o_{minc} = 1.25$, to avoid using interpolated cells for the outlier detection. The two thresholds for the point cloud segmentation were set to $h_g = 0.3m$ for cells classified as ground and $h_o = 0.1m$ for obstacle cells (see~\ref{sec:cloud_segmentation}). The confidence point count scaling factor was $s = 20$, so that a ground cell must contain at least 20 points to reach a perfect confidence rating of 1 (see~\ref{eq:cur_confidence}). The outlier tolerance was set to $o_{t} = 0.1m$, so points that were occluded by less than this height are still considered valid. The distance where the patch size changes from $3 \times 3$ to $5 \times 5$ was $d_{ps} = 20m$. The expected point distance was $d_{pv} = 0.4\degree$ (see~\ref{eq:expected_points}) which is based on the pessimistic assumption that at least half of the beams of the HDL64e produced a return at 10Hz. The point count threshold to use the cell's variance instead of the patch's was $v_{np} = 10$. Our implementation uses the Robot Operating System (ROS) and the Grid Map library~\cite{fankhauser16}. We evaluated GroundGrid's ground segmentation performance with the SemanticKITTI dataset. We define the labels \textit{lane-marking}, \textit{other-ground}, \textit{parking}, \textit{road}, \textit{sidewalk}, and \textit{terrain} as ground. The labels \textit{unlabeled} and \textit{outlier} are ignored in the error metrics since they do not belong to either class. Points labeled \textit{vegetation} are also ignored in the error metrics. Due to the way the \textit{vegetation} class is labeled in the SemanticKITTI dataset it is impossible to say if the majority of point belonging to this class belong to the ground surface or not. Hence we decided to follow the example of Lee et al.~\cite{lee22} and exclude this label from the error metrics (it is still included in the point clouds). All other labels are evaluated as non-ground. 
\subsection{Ground Segmentation Evaluation}
Fig.~\ref{tab:qualitative_semkitti} shows qualitative results of the SemanticKITTI dataset. From top to bottom we compare the provided ground truth with the results of the methods \textit{Patchwork++}~\cite{lee22}, \textit{JPC}~\cite{shen21} and \textit{GndNet}~\cite{paigwar20}, and \textit{GroundGrid} (ours). The first scenario \textit{a)} is cloud 40 of sequence \textit{01}. One notable difficulty in this scenario is the ditch located in the right part of the image. GroundGrid is the only method segments the points located in and around this ditch correctly while other methods over-segment the outer slope. The same applies to the points around the curbs where the other method display over-segmentation. Some methods fail to correctly segment the points on the opposite lane left to the vehicle because most laser beams are blocked by the guardrail. The distant guardrail in the lower right part is correctly segmented by all methods except JPC which slightly over-segments it and ours which partly under-segments it.
\begin{figure*}
  \centering
   \begin{tabular}{c<{\hspace{-.75em}}c<{\hspace{-.75em}}c<{\hspace{-.75em}}c}
   Urban &
   \includegraphics[width=2.1in,valign=c]{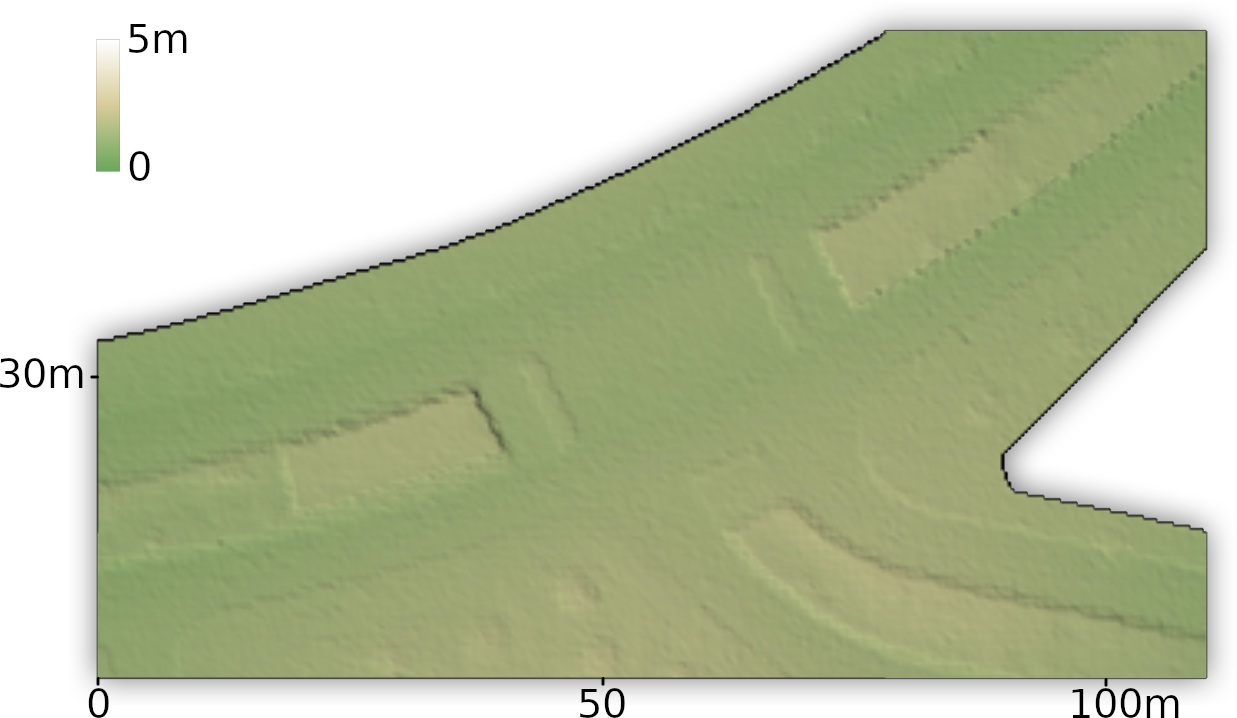} &
   \includegraphics[width=2.1in,valign=c]{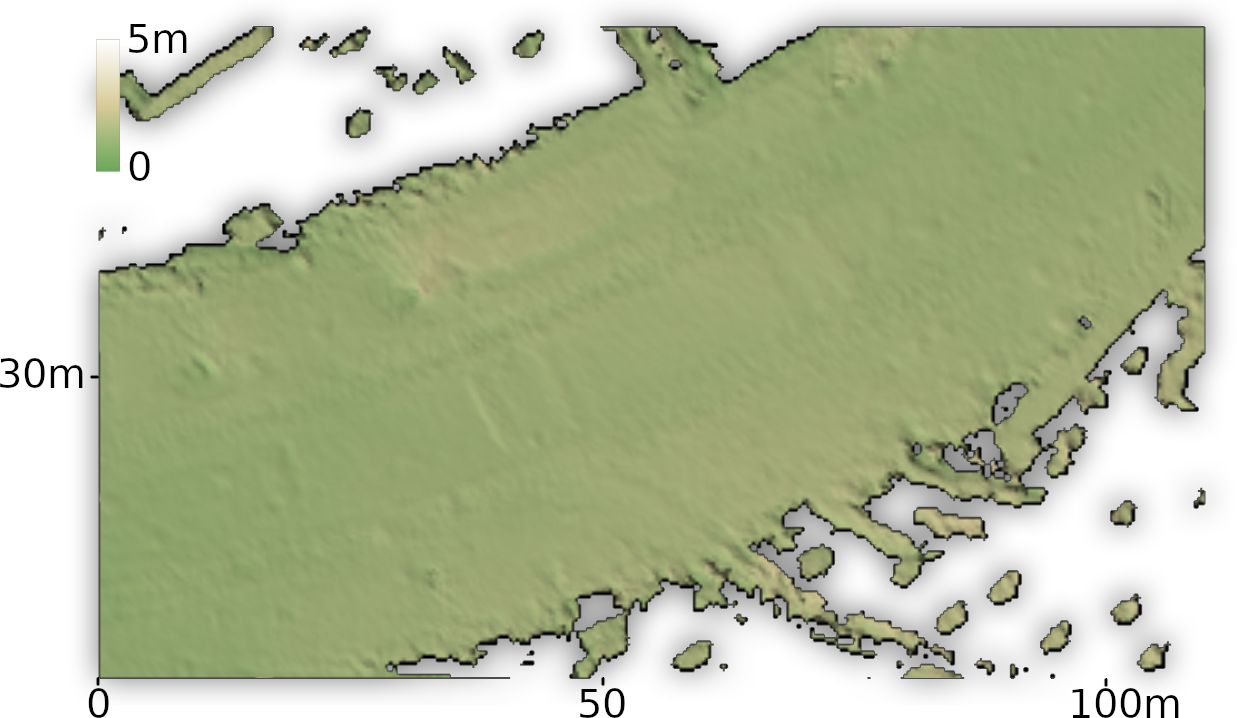} &
   \includegraphics[width=2.1in,valign=c]{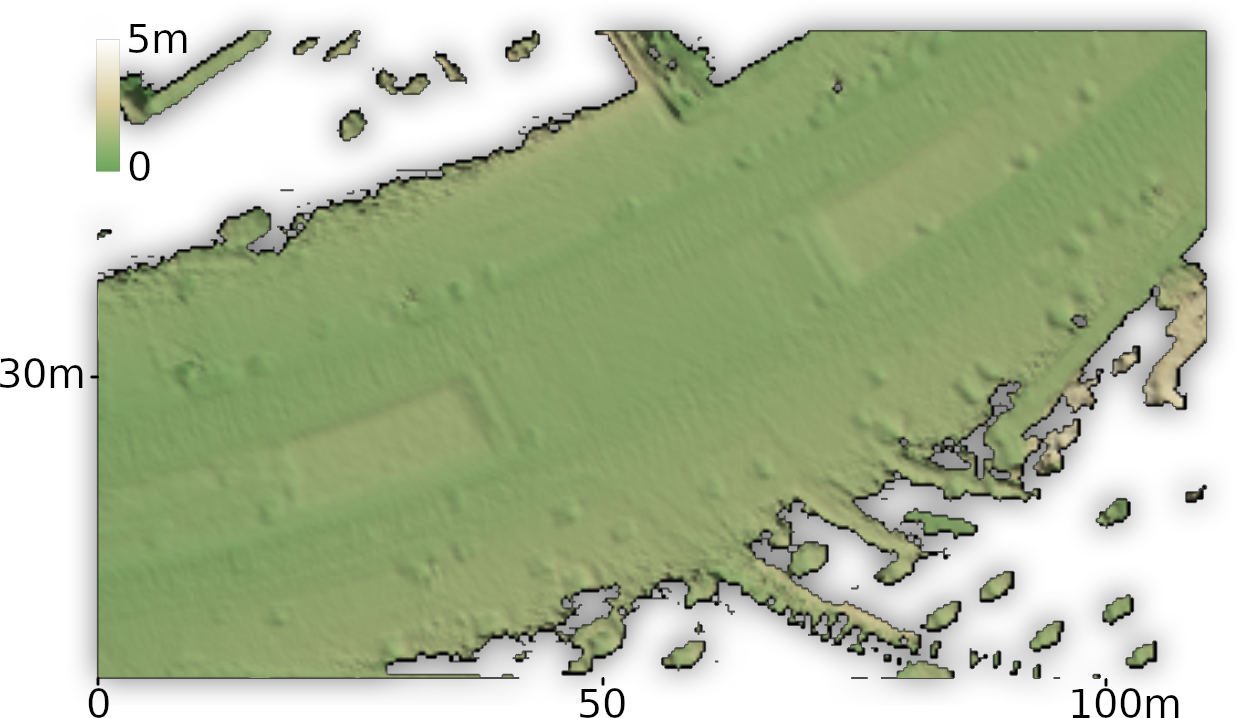}\\
      Hill &
   \includegraphics[width=2.1in,valign=c]{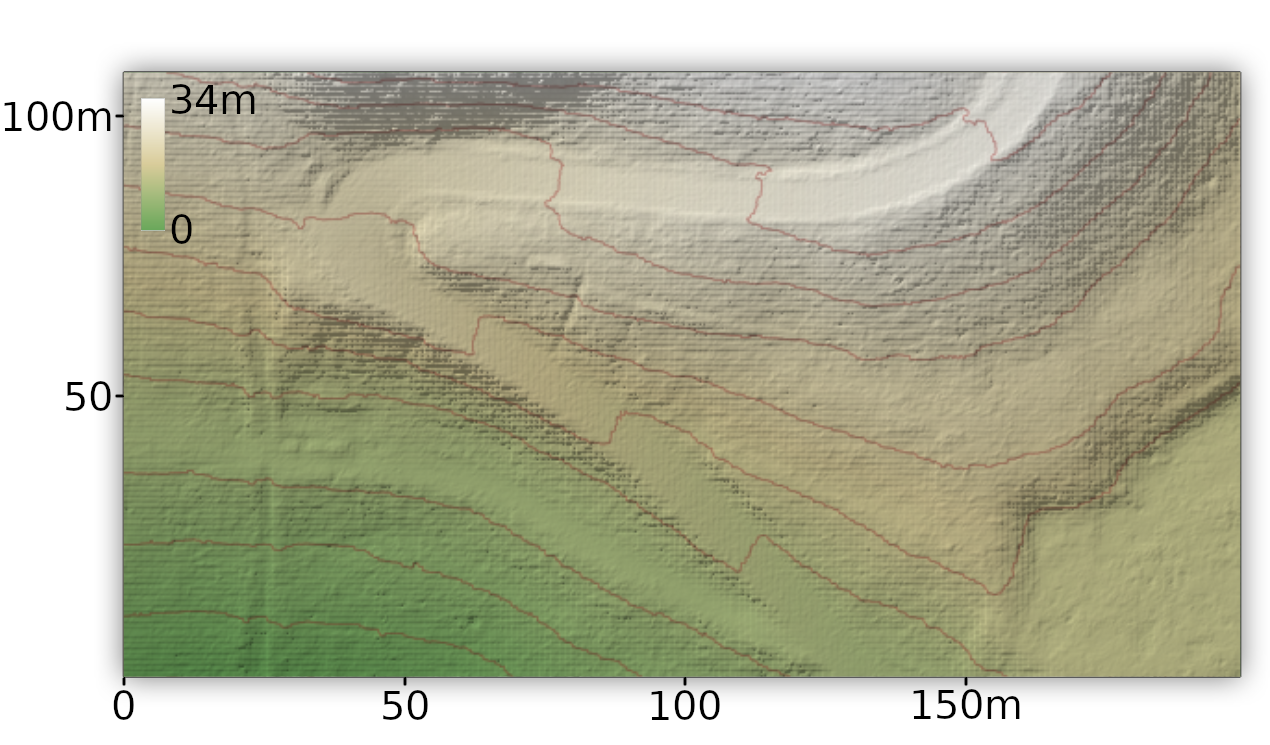} &
   \includegraphics[width=2.1in,valign=c]{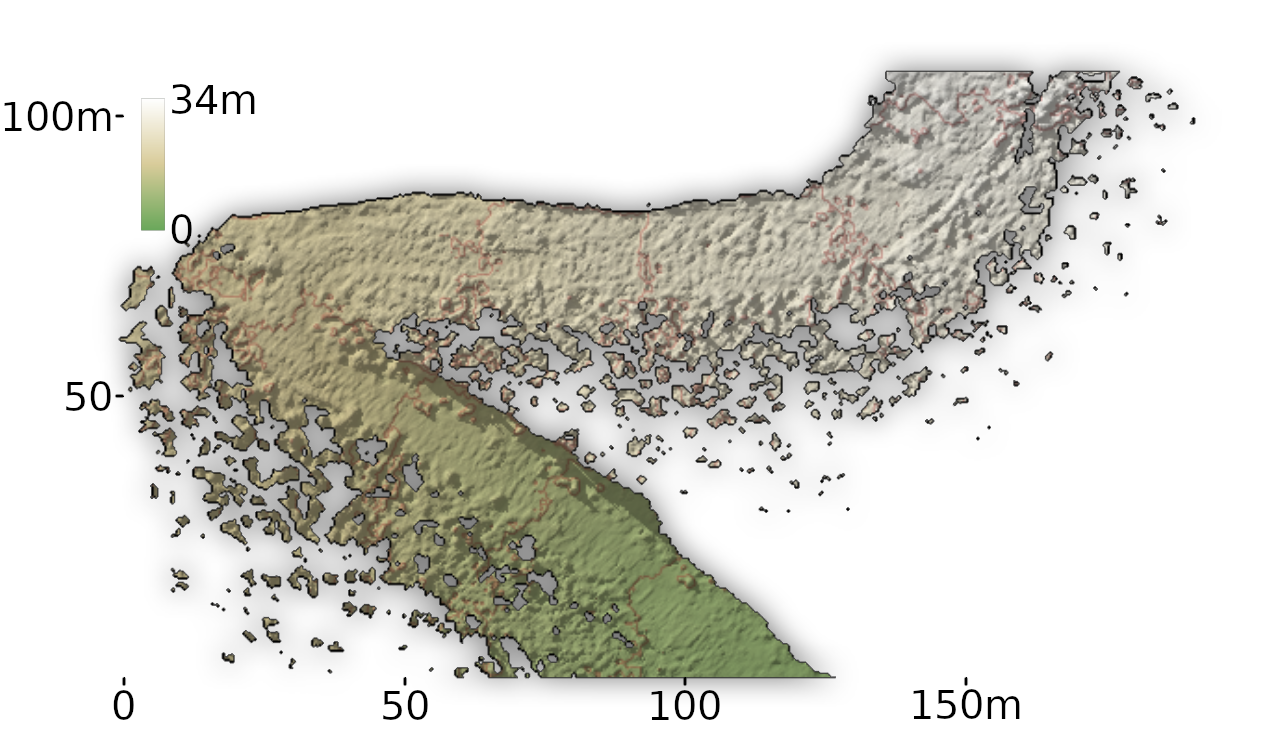} &
   \includegraphics[width=2.1in,valign=c]{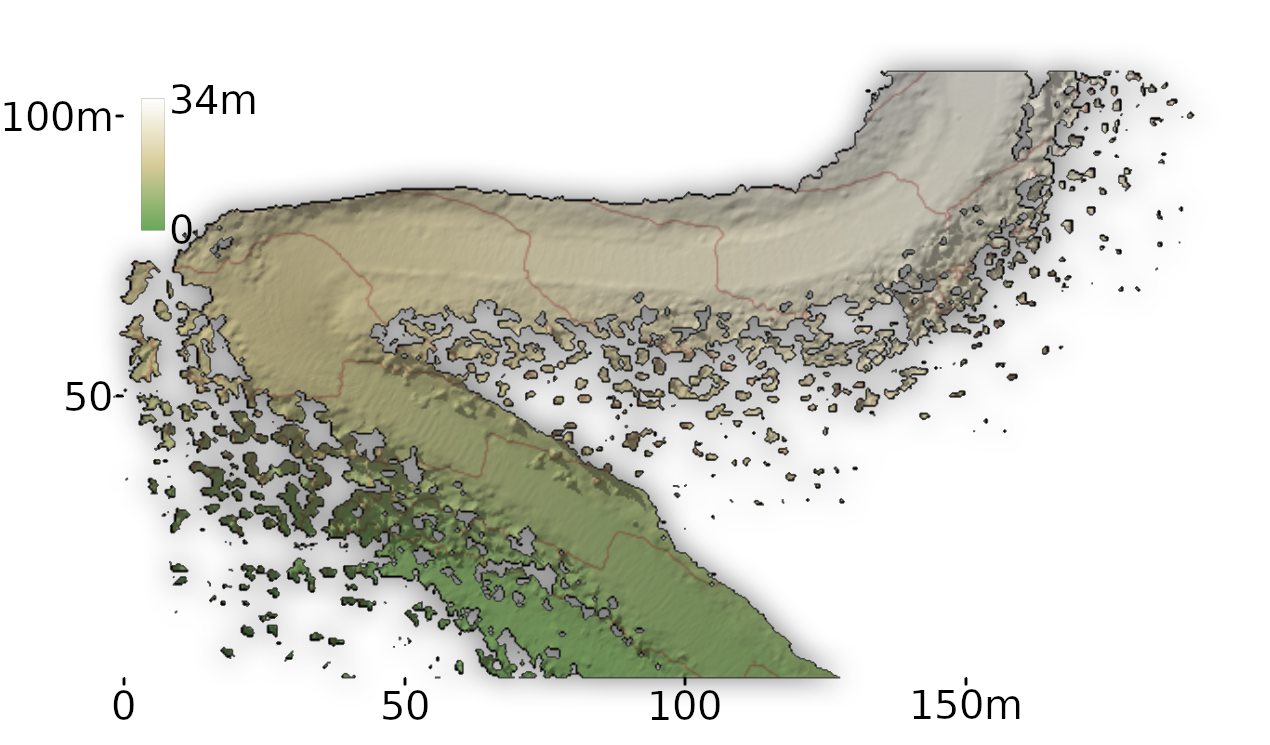}\\
   & Ground Truth~\cite{fis} & GndNet~\cite{paigwar20} & Ours\\
   \end{tabular}
      \caption{Terrain estimation qualitative evaluation}
   \label{tab:qualitative_terrain}
\end{figure*}
In \textit{b)} (seq. 09, cloud 1590) the methods have to deal with significant elevation differences and a steep slope in the driving direction. All methods over-segment parts of the road surface in this scenario. Patchwork++ performs well in lower streets towards the bottom but struggles with the higher elevated parts in the upper half of the image. GndNet struggles with the smaller streets to the sides. JPC over-segments the lower small street and the area around the curbs. GroundGrid manages to correctly segment almost all of the street surface but some over-segmentation can be found at the very top where a part of the curb is wrongly classified. Scenario \textit{c)} is a detailed view of a difficult segmentation situation (seq. 10, cloud 164). The non-continuous ground elevation changes combined with vegetation and tilted surfaces make the ground segmentation task especially challenging. Patchwork++ tends to over-segment which is visible in the sloped area to the center right and in the area in the back. There is also under-segmentation present at the wall to the center-back and right. GndNet severely struggles with over-segmentation in the sloped area to the center right while heavily under-segmenting the cars and walls in the center back. JPC over-segments the area to the center right as well as the ground surface in the center back. GroundGrid also suffers from over-segmentation in this scenario albeit less strongly compared to the other methods: There is some over-segmentation in the center right area around the vegetation and near the wall that separates the area from the street level. This is also true for the walls in the back part but the elevated ground surface is still mostly correctly segmented.
For the quantitative evaluation, we chose the following error metrics:
\begin{equation}
    \scriptsize
    \text{Precision} = \frac{TP}{TP + FP} \quad \text{Recall} = \frac{TP}{TP + FN}
\end{equation}
\begin{equation}
    \scriptsize
    F_1 = \frac{2 \cdot TP}{2 \cdot TP + FP + FN} \quad IoU = \frac{TP}{TP + FP + FN}
\end{equation}
\begin{equation}
    \scriptsize
    \text{Accuracy} = \frac{TP + TN}{TP + TN + FP + FN}
\end{equation}
where \textit{TP} and \textit{FP} represent the count of true positive and false positive ground points and \textit{TN} and \textit{FN} the count of true and false negative ground points. Table~\ref{tab:ground_segmentation} shows the results for all methods on the SemanticKITTI dataset. GroundGrid exhibits the best performance in most metrics except Recall and Precision where Patchwork++ and GndNet produce higher averages. The high recall for Patchwork++ comes with a lower precision compared to the other methods which suggests a tendency for over-segmentation. This is contrasted by GndNet which scores the highest in the recall metric but suffers from severe under-segmentation. Keep in mind that this method was trained on this very same dataset so possible overfitting could influence the results.
\subsection{Terrain Estimation Evaluation}
For the terrain estimation evaluation, we used the publicly available Airborne Lidar Scanning (ALS) data published by the city of Berlin~\cite{fis}. The maximum absolute height error is verified to be smaller than $0.05m$, validated with 30 known reference points~\cite{fis}.
We preprocessed the point clouds by filtering non-ground points and rasterizing them onto a grid with $0.5m$ cell size to create rasterized elevation maps which we used to evaluate the elevation maps produced by the algorithms (see ground truth in Fig~\ref{tab:qualitative_terrain}). We evaluate two experiments in different settings: The first in an urban road setting with mostly flat terrain, and the second in a forest area driving downwards in sloped terrain. We generated the data with our autonomous test vehicle equipped with a Velodyne HDL-64e 64-beam LiDAR placed on the car's roof and an Applanix POS-LV GNSS for localization purposes. 
Fig.~\ref{tab:qualitative_terrain} shows visualizations (created using~\cite{morgan-wall23}) of some of the results of the terrain estimation experiments. This data was collected by averaging the terrain estimation output of the methods. Only pixels with a point density of at least $27 \frac{points}{m^2}$ were considered. This equals $3$ points per $0.33m^2$ area on average which we define as a lower threshold for a meaningful variance calculation. For comparison purposes, we display the ALS ground truth alongside the results. The visualization of the urban scenario shows a T-type crossing where three streets meet. Our vehicle traveled straight hence the third street is not covered by the terrain estimation methods. The lanes are divided by artificially raised ground covered with trees and lower vegetation except for the center part of the crossing where all lanes are connected. Trees can also be found on the side of the streets which raises the difficulty for the terrain estimation. The dividing elevated area between the lanes can be seen in the ground truth and GroundGrid's results. Whereas GndNet's results only faintly show the outline of this area. However, GroundGrid struggles to correctly interpolate the area under the trees on the side of the road and the dividing area. This is represented by lower circular areas in the ground estimation, visible as darker spots in the visualization. GndNet does not display these spots but does seem to smooth the elevation map too much leading to a loss of detail. The results of the hill scenario show the performance in the presence of a significant elevation difference. The pictured area covers a trajectory length of approx. 250m of the car's trajectory with an elevation difference of 23m between the top and bottom point which translates into an average inclination of 9.2\%. Due to the dense tree cover on the roadsides, the ground visibility was low except on the street. Paigwar's GndNet struggled in this challenging environment which is shown by the high amount of noise in the resulting terrain estimation. The sides of this narrow street are covered with trees and the terrain is rough and uneven. Since GndNet was trained on the SemanticKITTI dataset it has never seen this type of terrain before and struggles to generalize its learned segmentation to this scenario. This is contrasted by the result of GroundGrid which is able to correctly extract the street's surface from the point cloud data. Only in the part after the turn, some noise is visible on the side of the road. Since the parameters of our method were the same in all experiments it shows that GroundGrid generalizes well over various scenarios.
 \begin{table}[t]
  \centering
    \vspace{.5em}
   \caption{Surface estimation accuracy evaluation (rsme in m)}
   \begin{tabular}{l||cc}
       & {urban} & {hill} \\
       \midrule
     GndNet & 0.297 & 1.581 \\
     Ours & \textbf{0.196} & \textbf{0.488} \\
     \bottomrule
   \end{tabular}
   \label{tab:terrain_eval}
 \end{table}
The root mean square error as measured against the ground truth data is shown in tab.~\ref{tab:terrain_eval}. GroundGrid achieves the best performance in both scenarios. The results show the high difficulty of the hill scenario, especially GndNet is not able to provide an accurate terrain estimation. The underlying network does seem to be able to generalize to this type of terrain due to the lack of similar training data. However, the performance of our method is also considerably lower in the hill scenario which seems to be caused by the occlusion due to the forest area.
\subsection{Run-Time}
GroundGrid has a high run-time performance of \textbf{171Hz} (5.85ms per point cloud) measured on an Intel i5-13600k desktop system.
The implementation of the point cloud rasterization, outlier detection, and ground cell classification is multi-threaded using up to eight threads in our evaluation. The ground interpolation and point cloud segmentation are single-threaded.
\section{CONCLUSION}
In this article, we presented a method for the point cloud ground segmentation and terrain estimation task consisting of outlier filtering, grid map rasterization, variance-based ground detection and the retention of previous information. The experimental results demonstrate the capabilities of the algorithm which outperforms current state-of-art methods in point cloud ground segmentation and terrain estimation tasks. This was demonstrated quantitatively and qualitatively using the SemanticKITTI dataset as well as self-acquired data from scenarios. The run-time performance is very high which enables online operation in mobile robots. The code of our implementation is available as open source and we invite the reader to reproduce our results.
%%%%%%%%%%%%%%%%%%%%%%%%%%%%%%%%%%%%%%%%%%%%%%%%%%%%%%%%%%%%%%%%%%%%%%%%%%%%%%%%%%%%%
%%%%%%%%%%%%%%%%%%%%%%%%%%%%%%%%%%%%%%%%%%%%%%%%%%%%%%%%%%%%%%%%%%%%%%%%%%%%%%%%
%%%%%%%%%%%%%%%%%%%%%%%%%%%%%%%%%%%%%%%%%%%%%%%%%%%%%%%%%%%%%%%%%%%%%%%%%%%%%%%%
%%%%%%%%%%%%%%%%%%%%%%%%%%%%%%%%%%%%%%%%%%%%%%%%%%%%%%%%%%%%%%%%%%%%%%%%%%%%%%%%
%%%%%%%%%%%%%%%%%%%%%%%%%%%%%%%%%%%%%%%%%%%%%%%%%%%%%%%%%%%%%%%%%%%%%%%%%%%%%%%%

\end{document}